\documentclass[10pt, a4paper]{scrartcl}

\usepackage{latexsym}
\usepackage{amssymb}
\usepackage[nonamelimits]{amsmath}
\usepackage[authoryear, round]{natbib}
\usepackage{url}

\usepackage{color}
\definecolor{myred}{rgb}{0.5,0,0}
\definecolor{myblue}{rgb}{0,0,0.75}
\definecolor{mygreen}{rgb}{0,0.5,0}

\usepackage{ifpdf}

\ifpdf
   \usepackage[pdftex]{graphicx}
   \usepackage[pdftex,
               pdftitle={Confidence intervals for class prevalences under prior probability shift},
               pdfsubject={},
               pdfkeywords={},
               pdfauthor={Dirk Tasche},
               pdfstartview=FitR,
               breaklinks=true,
               colorlinks=true,
               citecolor=myred,
               linkcolor=myblue,
               urlcolor=mygreen]{hyperref}
\else
   \usepackage[dvips]{graphicx}
   \usepackage{hyperref}
   \usepackage{rotating}
\fi

\numberwithin{equation}{section}

\title{Confidence intervals for class prevalences under prior probability shift}

\author{%
Dirk Tasche\thanks{E-mail: dirk.tasche@gmx.net}}%\newline
\date{June 10, 2019}

\begin{document}

\maketitle

\begin{abstract}
Point estimation of class prevalences in the presence of data set shift
has been a popular research topic for more than two decades. Less attention
has been paid to the construction of confidence and prediction intervals
for estimates of class prevalences. One little considered question is
whether or not it is necessary for practical purposes to distinguish confidence and prediction
intervals. Another question so far not yet conclusively answered is 
whether or not the discriminatory power of the classifier or score at
the basis of an estimation method matters for the accuracy of the estimates
of the class prevalences. This paper presents a simulation study aimed at
shedding some light on these and other related questions.  
\end{abstract}

%%%
% New section
%%%

\section{Introduction}
\label{se:intro}

In a prevalence estimation problem, one is presented with a sample of unlabelled instances (the test sample) 
and is asked to estimate the distribution of the labels in the sample. If the problem sits in a binary
two-class context,
all instances belong to exactly one of two possible classes and, accordingly, can be labelled
either positive or negative. The distribution of the labels then is characterised by 
the prevalence (i.e.~proportion) of the positive labels (`class prevalence' for short) in the test sample. 
However, the labels are latent at estimation time such that the class prevalence 
cannot be determined by simple inspection of the labels. Instead 
the  class prevalence  can only be inferred from the features of the
instances in the sample, i.e.\ from observable covariates of the labels. The interrelationship
between features and labels must be learnt from a training sample of labelled instances in another step
before the class prevalence of the positive labels in the test sample can be estimated. 

This whole process is called `supervised prevalence estimation' \citep{barranquero2013study}, `quantification' 
\citep{forman2008quantifying}, `class distribution estimation' \citep{gonzalez2013class}
or `class prior estimation' \citep{duPlessisEtAl2017} in the literature.
See \citet{Gonzalez:2017:RQL:3145473.3117807} for a recent overview of the quantification problem and
approaches to deal with it. The emergence of further recent papers with new proposals 
of prevalence estimation methods suggests that 
the subject is still of high interest for both researchers and practitioners 
\citep{castano.analisis,  Keith&OConnor, Maletzke&etAl.Quantification, Vaz&Izbicki&Stern2019}.

A variety of different methods for prevalence point estimation has been proposed and a considerable number
of comparative studies for such methods has been published in the literature \citep{Gonzalez:2017:RQL:3145473.3117807}. 
But the question of how to construct confidence and prediction intervals for class prevalences
seems to have attracted less attention. \citet{hopkins2010method} routinely provided confidence intervals
for their estimates ``via standard bootstrapping procedures'', 
without commenting much on details of the procedures or
on any issues encountered with them. \citet{Keith&OConnor} proposed and compared a 
number of methods for constructing such
confidence intervals. Some of these methods involve Monte-Carlo simulation and some do not.
Also \citet{daughton2019constructing} proposed 
a new method for constructing bootstrap confidence intervals and compared its
results with the confidence intervals based on popular prevalence estimation methods.
\citet{Vaz&Izbicki&Stern2019} introduced the `ratio estimator' for
class prevalences and used its asymptotic properties for determining confidence intervals without
involving Monte-Carlo techniques.

This paper presents a simulation study that seeks to illustrate some observations from
these previous papers on confidence intervals for class prevalences in the binary case and to
provide answers to some questions begged in the papers:
\begin{itemize}
\item Would it be worthwhile to distinguish confidence and prediction intervals 
for class prevalences and deploy different methods for their estimation? This question is raised 
against the backdrop that for instance \citet{Keith&OConnor} talked about estimating confidence 
intervals but in fact constructed prediction intervals which are conceptionally different \citep{MeekerEtAl}.
\item Would it be worthwhile to base class prevalence estimation on more accurate classifiers? The 
background for this question are conflicting statements in the literature as to the benefit 
of using accurate classifiers for prevalence estimation. On the one hand, 
\citet[][p.~168]{forman2008quantifying} stated: 
``A major benefit of sophisticated methods for quantification 
is that a much less accurate classifier can be used to obtain reasonably precise quantification estimates. 
This enables some applications of machine learning to be deployed where otherwise the raw classification
accuracy would be unacceptable or the training effort too great.'' As an example for the 
opposite position, on the other hand, \citet[][p.~595]{barranquero2015quantification} commented
with respect to prevalence estimation: 
``We strongly believe that it is
also important for the learner to consider the classification
performance as well. Our claim is that this aspect is crucial to
ensure a minimum level of confidence for the deployed models.'' 
\item Which prevalence estimation methods show the best performance with respect to the construction of
as short as possible confidence intervals for class prevalences?
\item Do non-simulation approaches to the construction of confidence intervals for
class prevalences work?
\end{itemize}
In addition, this paper introduces two new methods for class prevalence estimation which are specifically 
designed for delivering as short as possible confidence intervals.

Deploying a simulation study for finding answers to the above questions has some advantages
compared to working with real-world data:
\begin{itemize}
\item The true class prevalences are known and can even be chosen with a view to facilitate obtaining
clear answers.
\item The setting of the study can be freely modified -- say with regard to samples sizes or accuracy
of the involved classifiers -- in order to more precisely investigate the topics in question.
\item In a simulation study, it is easy to apply an ablation approach to assess the relative impact
of factors that influence the performance of methods for estimating confidence intervals.
\item The results can be easily replicated.
\item Simulation studies are good for delivering counter-examples. A method performing poorly
in the study reported in this paper may be considered unlikely to perform much better in complex
real-world settings.
\end{itemize}
Naturally, these advantages are bought at the cost of accepting certain obvious drawbacks:
\begin{itemize}
\item Most findings of the study are suggestive and illustrative only. No firm conclusions
can be drawn from them.
\item Important features of the problem which only occur in real-world situations might be overlooked.
\item The prevalence estimation problem primarily is caused by data set shift. For capacity reasons,
the scope of the simulation study in this paper is restricted to prior probability shift\footnote{%
In the literature, prior probability shift is known under a number of different names, for instance
`target shift' \citep{Zhang:2013:TargetShift}, `global drift' \citep{hofer2013drift}, or `label shift'
\citep{pmlr-v80-lipton18a}. See \citet{MorenoTorres2012521} for a categorisation of types of data set shift.% 
}, a special type of 
data set shift.
\end{itemize}
With these qualifications in mind, the main findings of this paper can be summarised as follows:
\begin{itemize}
\item Extra efforts to construct prediction intervals instead of confidence intervals for class
prevalences appear to be unnecessary.
\item `Error Adjusted Bootstrapping' as proposed by \citet{daughton2019constructing} for the
construction of prevalence confidence or prediction intervals may fail in the presence of 
prior probability shift.
\item Deploying more accurate\footnote{%
In this paper, instead of accuracy also the term `discriminatory power' is used. Similarly, 
instead of `accurate' the adjective `powerful' is employed.} classifiers for class 
prevalence estimation results in shorter confidence 
intervals.
\item Compared to the other estimation methods considered in this paper, 
straight-forward `adjusted classify \& count` methods for prevalence estimation 
(\citealp{forman2008quantifying}, called `confusion matrix method' in \citealp{saerens2002adjusting})
without any further tuning produce the longest confidence intervals and hence,
given identical coverage, perform worst. Methods based on minimisation of the Hellinger
distance \citep[][with different numbers of bins]{gonzalez2013class}  produce much shorter confidence intervals,
but sometimes do not guarantee sufficient coverage. The maximum likelihood approach 
(with bootstrapping for the confidence intervals) and `adjusted probabilistic classify \& count` 
(\citealp{bella2010quantification}, called there `scaled probability average') appear
to stably produce the shortest confidence intervals among the methods considered in the paper.
\end{itemize}
The paper is organised as follows:
\begin{itemize}
\item Section \ref{sec:setting} `Setting of the simulation study' describes the conception
and technical details of the simulation study, including in sub-section \ref{sec:approaches} 
a list of the prevalence estimation methods in scope.
\item Section \ref{sec:num} `Results of the simulation study' provides some tables with results of the study
and comments on the results, in order to explore the questions stated above. Results in subsection~\ref{sec:NoSim}
show that certain standard non-simulation approaches cannot take into account estimation uncertainty
in the training sample and that bootstrap-based construction of confidence intervals could be used 
instead.
\item Section \ref{se:concl} `Conclusions' wraps up and closes the paper.
\item In Appendix \ref{sec:app} `Particulars for the implementation of the simulation study', the mathematical
details needed for coding the simulation study are listed.
\item In Appendix \ref{sec:appB} `Analysis of Error Adjusted Bootstrapping', the appropriateness
for prior probability shift of the approach proposed by \citet{daughton2019constructing} is investigated.
\end{itemize}
The calculations of the simulation study have been performed by making use of the statistical software R 
\citep{RSoftware}. The R-scripts utilised can be downloaded at URL \url{https://www.researchgate.net/profile/Dirk_Tasche}.

%%%
% New section
%%%

\section{Setting of the simulation study}
\label{sec:setting}

The set-up of the simulation study is intended to reflect the situation that occurs when 
a prevalence estimation problem as described in Section~\ref{se:intro} has to be solved:
\begin{itemize}
\item There is a \emph{training sample} 
$(x_{1, \mathrm{P}}, y_{1, \mathrm{P}}), \ldots,$ $(x_{m, \mathrm{P}}, y_{m, \mathrm{P}})$ 
of observations of features $x_{i, \mathrm{P}}$ and class labels $y_{i, \mathrm{P}} \in \{-1,\, 1\}$ 
for $m$ instances\footnote{%
Instances with label $-1$ belong to the negative class, instances with label $1$ belong to the positive class.}.
By assumption,
this sample was generated from a joined distribution $\mathrm{P}(X,Y)$ (the training population distribution) of the
feature random variable $X$ and the label (or class) random variable $Y$.
\item There is a \emph{test sample} $x_{1, \mathrm{Q}}, \ldots, x_{n, \mathrm{Q}}$ of 
observations of features $x_{i, \mathrm{Q}}$ for $n$ instances. By assumption,
each instance has a \emph{latent class label} $y_{i, \mathrm{Q}}\in \{-1,\, 1\}$, and both the features and the labels 
were generated from a joined distribution $\mathrm{Q}(X,Y)$ (the test population distribution) of the
feature random variable $X$ and the label random variable $Y$.
\end{itemize}
The \emph{prevalence estimation} or \emph{quantification problem}  
then is to estimate the prevalence $q = \mathrm{Q}[Y=1]$ of the
positive class labels in the test population. Of course, this is only a problem if there is data set shift, i.e.\ if
$\mathrm{P}(X, Y) \not= \mathrm{Q}(X, Y)$ and as a likely consequence $p= \mathrm{P}[Y=1] \not= q$.

This paper deals only with the situation where the training population distribution $\mathrm{P}(X,Y)$ and the
test population distribution $\mathrm{Q}(X,Y)$ are related by \emph{prior probability shift} which means in
mathematical terms that
\begin{equation}\label{eq:prior}
\mathrm{Q}[X\in A\,|\,Y=-1] = \mathrm{P}[X\in A\,|\,Y=-1] \quad \text{and}\quad
\mathrm{Q}[X\in A\,|\,Y=1] = \mathrm{P}[X\in A\,|\,Y=1],
\end{equation}
for all subsets $A$ of the feature space such that $\mathrm{P}[X\in A]$ and $\mathrm{Q}[X\in A]$ are 
well-defined.

\subsection{The model for the simulation study}
\label{sec:model}

The classical binormal model with equal variances  
fits well into the prior probability shift setting for prevalence estimation of this paper. 
\citet{kawakubo2016computationally} used it as part of their experiments for comparing the performance of
prevalence methods. Logistic regression is a natural and optimal approach to the estimation of the 
binormal model with equal variances \citep[Section~6.1,][]{Cramer2003}. Hence when 
logistic regression is used for the estimation of the model in the simulation study, 
there is no need to worry about the results being invalidated by the deployment of a sub-optimal
regression or classification technique.
The binormal model is specified by
defining the two class-conditional feature distributions $\mathrm{P}(X\,|\,Y=-1)$ and $\mathrm{P}(X\,|\,Y=1)$
respectively. 

\begin{subequations}
\textbf{Training population distribution.} 
Both class-conditional feature distributions are normal, with equal variances, i.e.
\begin{equation}\label{eq:CondNormal}
    \mathrm{P}(X\,|\,Y=-1)  = \mathcal{N}(\mu, \sigma^2),\qquad
    \mathrm{P}(X\,|\,Y=1)  = \mathcal{N}(\nu, \sigma^2), 
\end{equation}
with $\mu < \nu$ and $\sigma > 0$.

\textbf{Test population distribution.} 
Same as the training population distribution, with $\mathrm{P}$ replaced by $\mathrm{Q}$, in order to satisfy
the assumption \eqref{eq:prior} on prior probability shift between training and test times. 

For the sake of brevity, in the following the setting with \eqref{eq:CondNormal} for  
both the training and the test sample is referred to as `double' binormal setting. 

Given the class-conditional population distributions as specified in \eqref{eq:CondNormal},
the unconditional training and test population distributions can be represented as 
\begin{equation}
\begin{split}
\mathrm{P}(X) & = p\,\mathrm{P}(X\,|\,Y=1) + (1-p)\,\mathrm{P}(X\,|\,Y=-1) \qquad \text{and}\\
\mathrm{Q}(X) & = q\,\mathrm{P}(X\,|\,Y=1) + (1-q)\,\mathrm{Q}(X\,|\,Y=-1),
\end{split}
\end{equation}
with $p= \mathrm{P}[Y=1]$ and $q= \mathrm{Q}[Y=1]$ as parameters whose values in the course of the
simulation study are selected depending on the purposes of the specific numerical experiments.
\end{subequations}

\textbf{Control parameters.} For this paper's numerical experiments, the values for the
parametrisation of the model
are selected from the ranges specified in the following list:
\begin{itemize}
\item $p  \in \{0.33, \,0.5, 0.67\}$ is the prevalence of the positive class in the training population. 
\item $m \in \{100,\, \infty\}$ is the size of training sample. In the case $m=\infty$, the training sample is considered
identical with the training population and learning of the model is unnecessary. 
In the case of a finite training sample, the number of instances with positive 
labels is non-random in order to reflect the fact that for model development purposes a pre-defined stratification 
of the training sample might be desirable and can be achieved 
by under-sampling of the majority class or by over-sampling of the
minority class. $m^+$ then is the size of the training sub-sample with positive labels, and
$m^-$ is the size of the training sub-sample with negative labels. Hence it holds that
$m = m^+ + m^-$, $m^+= p\,m$ und $m^-= (1-p)\,m$ for finite $m$.
\item $q  \in \{0.05, \,0.2\}$ is the prevalence of the positive class in the test population.
\item $n \in \{50, 500\}$ is the size of the test sample. In the test sample, the number of instances
with positive labels is random.
\item The population distribution underlying the features of the negative-class training sub-sample
is always $\mathrm{P}(X\,|\,Y=-1)  = \mathcal{N}(\mu, \sigma^2)$ with $\mu = 0$ and $\sigma=1$.
The population distribution underlying the features of the positive-class training sub-sample
is $\mathrm{P}(X\,|\,Y=1)  = \mathcal{N}(\nu, \sigma^2)$ with $\nu \in \{1, 2.5\}$ and $\sigma=1$.
\item The population distribution underlying the features of the negative-class test sub-sample
is always $\mathrm{Q}(X\,|\,Y=-1)  = \mathcal{N}(\mu, \sigma^2)$ with $\mu = 0$ and $\sigma=1$.
The population distribution underlying the features of the positive-class test sub-sample
is $\mathrm{Q}(X\,|\,Y=-1)  = \mathcal{N}(\mu, \sigma^2)$ with $\nu \in \{1, 2.5\}$ and $\sigma=1$.
\item The number of simulation runs in all of the experiments is $n_{\text{sim}}=100$, 
i.e.~$n_{\text{sim}}$-times
a training sample and a test sample as specified above are generated and subjected to some estimation
procedures.
\item The number of bootstrap iterations where needed in any of the interval estimation 
procedures is always $R = 999$ \citep{DavisonHinkley}.
\item All confidence and prediction intervals are constructed at $\alpha = 90\%$ confidence level.
\end{itemize}
\begin{figure}[tb!p]
{\small\caption{Receiver Operating Characteristics for the high power ($\nu=2.5$) and
low power ($\nu=1$) simulation scenarios.}\label{fig:ROC}}
\begin{center}
\ifpdf
	\resizebox{\height}{9cm}{\includegraphics[width=10cm]{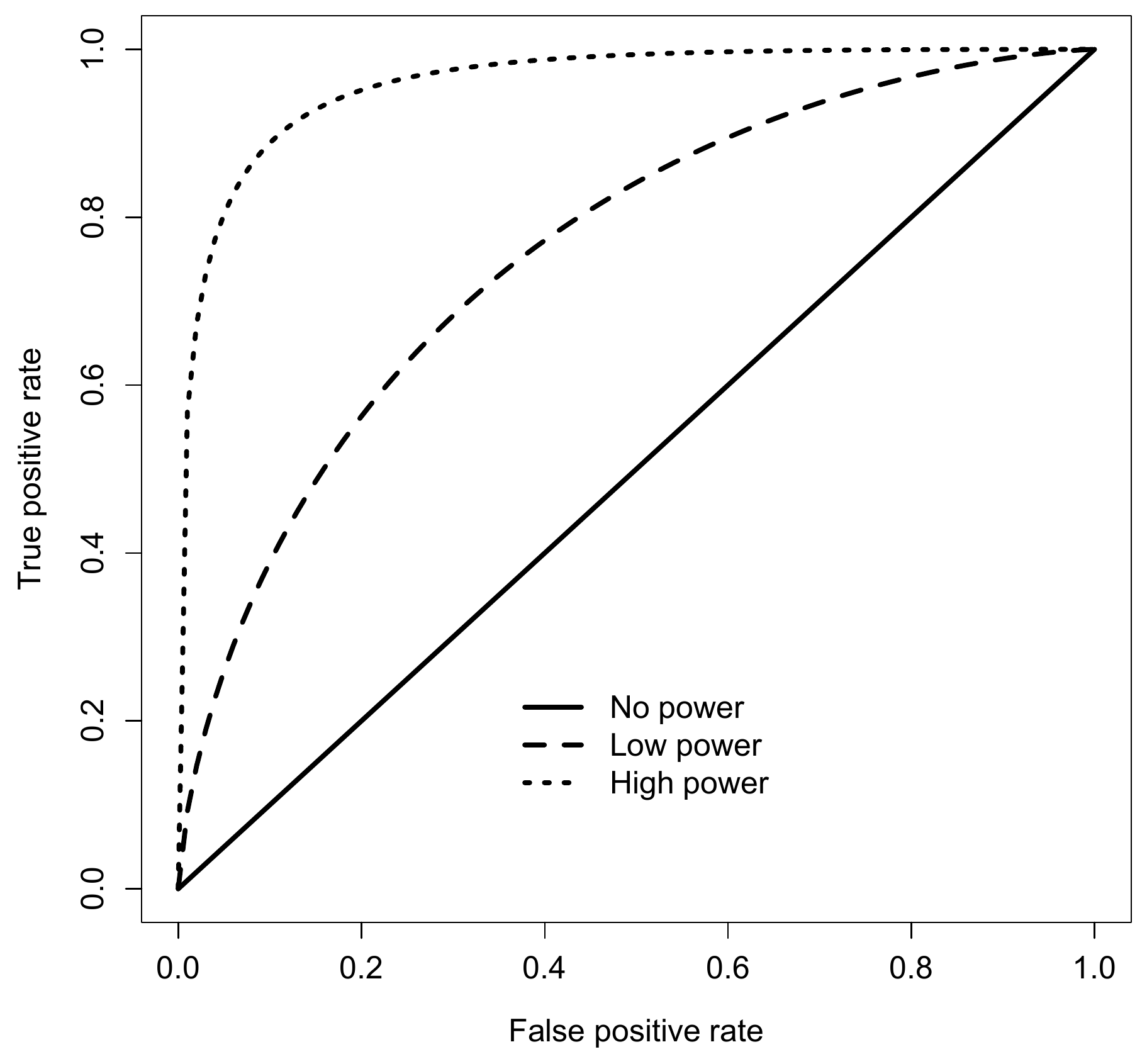}}
%\else
%\begin{turn}{270}
%\resizebox{\height}{15.0cm}{\includegraphics[width=8cm]{RatingProfiles.eps}}
%\end{turn}
\fi
\end{center}
\end{figure}

Choosing $\nu =1$ 
in one of the following simulation experiments will reflect a situation where no accurate classifier 
can be found, as it is suggested by the fact that then the AUC (area under the curve) of the feature $X$ taken
as a soft classifier is\footnote{%
$\Phi$ denotes the standard normal distribution function.} $\Phi\left(\frac{\nu-\mu}
{\sigma\,\sqrt{2}}\right) = 76.02\%$. In the case $\nu=2.5$ the
same soft classifier $X$ is very accurate with an AUC of $\Phi\left(\frac{\nu-\mu}{\sigma\,\sqrt{2}}\right) = 96.15\%$.
The different performance of the classifier depending on the value of parameter $\nu$ is also demonstrated
in Figure~\ref{fig:ROC}
by the ROCs (receiver operating characteristics) corresponding to the two values $\nu=1$ (`low power') and 
$\nu=2.5$ (`high power').

For the sake of completeness, it is also noted that 
the feature-conditional class probability $\mathrm{P}[Y=1\,|\,X]$ under the training population distribution is given by
\begin{subequations}
\begin{equation}\label{eq:trainCondProb}
    \mathrm{P}[Y=1\,|\,X](x) \  = \ \frac{1}{1 + \exp(a\,x + b)}, \quad x \in \mathbb{R},
\end{equation}
with $a = \frac{\mu-\nu}{\sigma^2} < 0$ and $b = \frac{\nu^2-\mu^2}{2\,\sigma^2} + 
\log\left(\frac{1-\mathrm{P}[Y=1]}{\mathrm{P}[Y=1]}\right)$.  
For the density ratio $R$ under both the training and test population distributions one obtains\footnote{%
$\phi_{\gamma, \tau}$ denotes the density function of the one-dimensional normal distribution with 
mean $\gamma$ and standard deviation $\tau$.}
\begin{equation}\label{eq:ratioBinorm}
R(x) \ = \ \frac{\phi_{\nu, \sigma}(x)}{\phi_{\mu, \sigma}(x)}\ = \ \exp\left(x\,\tfrac{\nu-\mu}{\sigma^2} + \tfrac{\mu^2-\nu^2}{2\,\sigma^2}\right),
\quad x \in \mathbb{R}.
\end{equation}
\end{subequations}

%%%
% New section
%%%

\subsection{Methods for prevalence estimation considered in this paper}
\label{sec:approaches}

The following criteria have been applied for the selection of the methods deployed in the simulation study:
\begin{itemize}
\item The methods must be Fisher consistent in the sense of \citet{tasche2017fisher}. This criterion excludes for instance
`classify \& count' \citep{forman2008quantifying}, the `Q-measure' approach  \citep{barranquero2013study} and the
distance-minimisation approaches based on the Inner Product, Kumar-Hassebrook, Cosine, and Harmonic Mean
distances mentioned in \cite{Maletzke&etAl.Quantification}.
\item The methods should enjoy some popularity in the literature.
\item Two new methods based on already established methods and designed to minimise the lengths of confidence 
intervals are introduced and tested.
\end{itemize}
According to these criteria the following prevalence estimation methods have been included in
the simulation study:
\begin{itemize}
\item ACC50: Adjusted Classify \& Count (ACC: \citealp{buck1966comparison, saerens2002adjusting, forman2008quantifying}), 
based
on the Bayes classifier that minimises accuracy. `50' because if the Bayes classifier is represented by means 
of the posterior probability of the positive class and a threshold, the threshold has to be 50\%.  
\item ACCp:  Adjusted Classify \& Count, based
on the Bayes classifier that maximises the difference of TPR (true positive rate) and 
FPR (false positive rate). `p' because if the Bayes classifier is represented by means 
of the posterior probability of the positive class and a threshold, the threshold needs to be $p$, the 
a priori probability (or prevalence) of the positive class in the training population. ACCp was called 
`method max' in \citet{forman2008quantifying}.
\item ACCv: New version of ACC where the threshold for the classifier is selected in such a way that the
variance of the prevalence estimates is minimised among all ACC-type estimators based on classifiers represented
by means of the posterior probability of the positive class and some threshold. 
\item MS: `Median sweep' as proposed by \citet{forman2008quantifying}.
\item APCC: `Adjusted probabilistic classify \& count' (\citealp{bella2010quantification}, 
there called `scaled probability average'). 
\item APCCv: New version of APCC where the a priori positive class probability parameter in the  
posterior positive class probability is selected in such a way that the
variance of the prevalence estimates is minimised among all APCC-type estimators based on 
posterior positive class probabilities where the a priori positive class probability parameter varies between
0 and 1.
\item H4: Hellinger distance approach with 4 bins \citep{gonzalez2013class, castano.analisis}.   
\item H8: Hellinger distance approach with 8 bins \citep{gonzalez2013class, castano.analisis}.
\item Energy: Energy distance approach    \citep{kawakubo2016computationally, castano.analisis}.
\item MLinf / MLboot: ML is the maximum likelihood approach to prevalence estimation \citep{peters1976numerical}.
Note that the EM (expectation maximisation) approach of \citet{saerens2002adjusting} is one way to 
implement ML. `MLinf' refers to construction of the prevalence confidence interval based on 
the asymptotic normality of the ML estimator (using the Fisher information for the variance). `MLboot'
refers to construction of the prevalence confidence interval solely based on bootstrap sampling.
\end{itemize}
For the readers' convenience, the particulars needed to implement the
methods in this list are presented in Appendix~\ref{sec:app}. 
Note that ACC50, ACCp, ACCv, APCC und APCCv are all special cases of the `ratio estimator'
discussed in \citet{Vaz&Izbicki&Stern2019}.

On the basis of the general asymptotic efficiency of maximum likelihood
estimators \citep[Theorem 10.1.12,][]{Casella&Berger},  
the maximum likelihood approach for class prevalences 
is a promising approach for achieving minimum confidence intervals lengths.
In addition, the ML approach may be considered a representative of the class of 
entropy-related estimators and, as such, is closely related to the Tops\o{}e approach 
which was found to perform very well in \citet{Maletzke&etAl.Quantification}.

\subsection{Calculations performed in the simulation study}
\label{sec:calc}

The calculations performed as part of the simulation study serve the purpose of providing
facts for answers to the questions listed in Section~\ref{se:intro} `Introduction'.

\textbf{Calculations for constructing confidence intervals.} 
Iterate $n_{\text{sim}}$ times the following 
steps:
\begin{enumerate}
\item Create the training sample: Simulate $m^+$ times from $\mathrm{P}(X\,|\,Y=1)  = \mathcal{N}(\nu, \sigma^2)$ features
$x_{1, \mathrm{P}^+}$, $\ldots$, $x_{{m^+}, \mathrm{P}^+}$ of positive instances and $m^-$ times from
$\mathrm{P}(X\,|\,Y=-1)  = \mathcal{N}(\mu, \sigma^2)$
features $x_{{1}, \mathrm{P}^-}$, $\ldots$, $x_{{m^-}, \mathrm{P}^-}$ of negative instances. 
\item Create the test sample: Simulate the number $N^+$ of positive instances as a binomial random variable
with size $n$ and success probability $q$. Then simulate $N^+$ times from 
$\mathrm{Q}(X\,|\,Y=1)  = \mathcal{N}(\nu, \sigma^2)$ features
$x_{1, \mathrm{Q}}$, $\ldots$, $x_{{N^+}, \mathrm{Q}}$ of positive instances and $N^- = n - N^+$ times 
from $\mathrm{Q}(X\,|\,Y=-1)  = \mathcal{N}(\mu, \sigma^2)$
features $x_{{N^+ +1}, \mathrm{Q}}$, $\ldots$, $x_{{n}, \mathrm{Q}}$ of negative instances. The information
of whether a feature $x_{i, \mathrm{Q}}$ was sampled from $\mathrm{Q}(X\,|\,Y=1)$ or from
$\mathrm{Q}(X\,|\,Y=-1)$ is assumed to be unknown in the estimation step. Therefore, the gnerated
features are combined in a single sample $x_{1, \mathrm{Q}}$, $\ldots$, $x_{{n}, \mathrm{Q}}$.
\item Iterate $R$ times the bootstrap procedure: Generate by stratified sampling with replications 
bootstrap samples $x'_{1, \mathrm{P}^+}$, $\ldots$, $x'_{{m^+}, \mathrm{P}^+}$ of features of positive 
instances,
$x'_{{1}, \mathrm{P}^-}$, $\ldots$, $x'_{{m^-}, \mathrm{P}^-}$ of features of negative instances from
the training subsamples, and 
$x'_{1, \mathrm{Q}}$, $\ldots$, $x'_{n, \mathrm{Q}}$ of features with unknown labels from the test sample. Calculate, based
on the three resulting bootstrap samples,  estimates of the positive class prevalence in the test population  
according to all the estimation methods listed in Section~\ref{sec:approaches}.
\item For each estimation method, the bootstrap procedure from the previous step creates a sample of
$R$ estimates of the positive class prevalence. Based on this sample of $R$ estimates, 
construct confidence intervals at level $\alpha$ for the positive class prevalence in the test population.
\end{enumerate}
\textbf{Tabulated results of the simulation algorithm for confidence intervals.}\ 
\begin{itemize}
\item For each estimation method, $n_{\text{sim}}$ estimates of the positive class prevalence 
are calculated. 
From this set of estimates, the following summary results are derived and tabulated:
\begin{itemize}
\item The average of the prevalence estimates.
\item The average absolute deviation of the prevalence estimates from the true prevalence parameter.
\item The percentage of simulation runs with failed prevalence estimates.
\item The percentage of estimates equal to 0 or 1.
\end{itemize}
\item For each estimation method, $n_{\text{sim}}$ confidence intervals at level $\alpha$ for 
the positive class prevalence are produced. From this set of confidence intervals, the following
summary results are derived and tabulated:
\begin{itemize}
\item The average length of the confidence intervals.
\item The percentage of confidence intervals that contain the
true prevalence parameter (coverage rate).
\end{itemize}
\end{itemize}
For the construction of the bootstrap confidence intervals in Step~4 of the list of calculations, the method
`perc' \citep[][Section~5.3.1]{DavisonHinkley} of the function boot.ci 
of the R-package `boot' is used. More accurate methods for bootstrap confidence
intervals are available, but these tend to require more computational time and to be less robust. Given that 
the performance of `perc' in the setting of this simulation study can be controlled via checking the coverage rates,
the loss in performance seems tolerable. In the cases where calculations have resulted in coverage rates of less than 
$\alpha$ the calculations have been repeated with the `bca` method \citep[][Section~5.3.2]{DavisonHinkley} 
of boot.ci in order to confirm the results.

Step~1 of the calculations can be omitted in the case $m = \infty$, i.e.\ when the training sample is identical
with the training population distribution. However, in this case some quantities of relevance for 
the estimates have to be pre-calculated before the entrance into the loop for the $n_{\mathrm{sim}}$ simulation runs. The
details for these pre-calculations are provided in Appendix~\ref{sec:app}. Also in the case $m = \infty$, for
the prevalence estimation methods ACC50, ACCp, ACCv, APCC und APCCv,
the bootstrap confidence intervals for the prevalences are replaced by ``conservative binomial intervals'' 
\citep[][Section~6.2.2]{MeekerEtAl}, computed with the `exact' method of the R-function binconf. Moreover, as explained in 
Section~\ref{sec:approaches}, in the case $m = \infty$ method MLinf is applied instead of MLboot for the 
construction of the maximum likelihood confidence interval.

As mentioned in Section~\ref{se:intro} `Introduction', one of the purposes of the simulation study is to 
illustrate the differences between confidence and prediction intervals. Conceptionally, the difference may be
described by their definitions as given in \citet{MeekerEtAl}\footnote{%
`$1-\alpha$' as used by \citet{MeekerEtAl} corresponds to `$\alpha$' as used in this paper.
}:
\begin{itemize}
\item ``A $100 (1-\alpha)\%$  \emph{confidence interval} for an unknown quantity $\theta$ may be formally characterized
as follows: If one repeatedly calculates such intervals from many independent random samples, $100 (1-\alpha)\%$
of the intervals would, in the long run, correctly include the actual value $\theta$. Equivalently, one would,
in the long run, be correct $100 (1-\alpha)\%$ of the time in claiming that the actual value of $\theta$
is contained within the confidence interval.'' \citep[][Section~2.2.5]{MeekerEtAl}
\item ``If from many independent pairs of random samples, a $100 (1-\alpha)\%$ \emph{prediction interval} is 
computed from the data of the
first sample to contain the value(s) of the second sample, $100 (1-\alpha)\%$ of the intervals would, in the long run, 
correctly bracket the future value(s). Equivalently, one would, in the long run, be correct $100 (1-\alpha)\%$ of the time in 
claiming that the future value(s) will be contained within the prediction interval.''
\citep[][Section~2.3.6]{MeekerEtAl} 
\end{itemize}
In order to construct prediction intervals instead of confidence intervals in the simulation runs, Step~4 of the
calculations is modified as follows:
\begin{enumerate}
\item[4')]  For each estimation method, the bootstrap procedure from the previous step creates a sample of
$R$ estimates of the positive class prevalence. 
For each estimate, generate a virtual number of realisations of positive instances by
simulating an inpendent binomial variable with size $n$ and success probability given by the estimate. Divide these
virtual numbers by $n$ to obtain (for each estimation method) a sample of relative frequencies of
positive labels.
Based on this additional size-$R$ sample  of relative frequencies, construct prediction intervals at level $\alpha$
for the percentage of instances with positive labels in the test sample.
\end{enumerate}
As in the case of the construction of confidence intervals, for the construction of the prediction intervals
again the method `perc' of the function boot.ci of the R-package `boot' is deployed.

\begin{samepage}
\textbf{Tabulated results of the simulation algorithm for prediction intervals.}\ 
\begin{itemize}
\item For each estimation method, $n_{\text{sim}}$ virtual relative frequencies of positive
labels in the test sample are simulated under the assumption that the estimated positive class
prevalence equals the true prevalence. 
From this set of frequencies, the following summary results are derived and tabulated:
\begin{itemize}
\item The average of the virtual relative frequencies.
\item The average absolute deviation of the virtual relative frequencies from the true prevalence parameter.
\item The percentage of simulation runs with failed prevalence estimates and hence also failed simulations
of virtural relative frequencies of positive labels.
\item The percentage of virtual relative frequencies equal to 0 or 1.
\end{itemize}
\item For each estimation method, $n_{\text{sim}}$ prediction intervals at level $\alpha$ for 
the realised relative frequencies of positive labels are produced. From this set of prediction intervals, the following
summary results are derived and tabulated:
\begin{itemize}
\item The average length of the prediction intervals.
\item The percentage of prediction intervals that contain the
true relative frequencies of positive labels (coverage rate).
\end{itemize}
\end{itemize}
\end{samepage}

\begin{table}[b!p]
{\small \caption{Explanation of the row names in the result tables of Section~\ref{sec:num}.}\label{tab:rows}
\vspace{-1ex}
\begin{center}
\begin{tabular}{|l|l|}\hline
Row name & Explanation \\ \hline \hline
`Av~prev' &  Average of the prevalence estimates (for confidence intervals)\\ \hline
`Av~freq' & Average of the relative frequencies of simulated positive class labels (for prediction intervals)\\ \hline
`Av~abs~dev' & Average of the absolute deviation of the prevalence estimates\\ 
 & or the simulated relative frequencies from the true prevalence\\ \hline
`Perc fail est' & Percentage of simulation runs with failed prevalence estimates\\ \hline
`Av~int~length'& Average of the confidence or prediction interval lengths\\ \hline
`Coverage' & Percentage of confidence intervals containing the
true prevalence  or \\
 & of prediction intervals containing the true realised relative frequencies of positive labels \\ \hline
`Perc 0 or 1'& Percentage of prevalence estimates or simulated fequencies with value $\le 10^{-7}$ or $\ge 1 - 10^{-7}$\\ \hline
\end{tabular} 
\end{center} }
\end{table}

%%%
% New section
%%%

\section{Results of the simulation study}
\label{sec:num}

All simulation procedures are performed with parameter setting $n_{\text{sim}} = 100$, $Rseed=17$ and $R = 999$ 
(see Section~\ref{sec:model} for the complete list of control parameters). At each table in the following, 
the values selected for the remaining control parameters are listed in the captions or within the table bodies.

In all the simulation procedures run for this paper, the R-boot.ci method for determining the statistical intervals (both confidence and prediction) has been
the method `perc'. In cases where the coverage found with `perc' is significantly lower than 90\% (for $n_{\text{sim}} = 100$ 
at 5\% significance level this means lower than 85\%), the calculation has been repeated with the R-boot.ci method `bca'
for confirmation or correction.

The naming of the table rows and table columns has been standardized. Unless mentioned otherwise, 
the columns always display results
for all or some of the prevalence estimation methods listed in Section~\ref{sec:approaches}. Short explanations of the
meaning of the row names are given in Table \ref{tab:rows}. A more detailed explanation of the row names
can be found in Section~\ref{sec:calc}.

\subsection{Prediction vs.\ confidence intervals}
\label{sec:predVSconf}

In the simulation study performed for this paper, the values of the true positive class prevalences of the 
test samples -- understood in the sense of the a priori positive class prevalences of the populations from
which the samples were generated (see Section~\ref{sec:setting}) -- are always known. In contrast, when one is
working with real-world data sets, there is no way to know with certainty the true positive class prevalences of the 
test samples. Inevitably, therefore, in studies of prevalence estimation methods on real-world data sets, the
performance has to be measured by comparison between the estimates and the relative frequencies of 
the positive labels observed in the test samples.

This was stated explicitly, for instance, in \citet{Keith&OConnor}. The authors said in the section 
`Problem definition' of the paper that they estimated `prevalence
confidence intervals' with the property that ``$(1-\alpha)\%$ of the predicted intervals
ought to contain the true value $\theta^\ast$''. For this purpose, 
\citeauthor{Keith&OConnor} defined the `true value' as follows: 
``For each group $\mathcal{D}$, let $\theta^\ast \equiv \left(1/n\right) \sum_i^n y_i$ 
be the true proportion of positive labels (where $n = |\mathcal{D}|$).'' As `group' was used by
\citeauthor{Keith&OConnor} as equivalent to sample and $y_i$ was 1 for positive labels and 0 otherwise, 
it is clear that \citeauthor{Keith&OConnor} estimated rather 
prediction intervals than confidence intervals (see Section~\ref{sec:calc} for the definitions of both
types of intervals). 

Hence, would it be worthwhile to distinguish confidence and prediction intervals 
for class prevalences and deploy different methods for their estimation, as has been asked in 
Section~\ref{se:intro}? 

By assumption (see Section~\ref{sec:setting}), the test sample $x_{1, \mathrm{Q}}, \ldots, x_{n, \mathrm{Q}}$ 
is interpreted as the feature components of independent, identically distributed random variables
$(X_{1, \mathrm{Q}}, Y_{1, \mathrm{Q}})$,  $\ldots$,  $(X_{n, \mathrm{Q}}, Y_{n, \mathrm{Q}})$. While
the positive class prevalence in the test population is given by the constant $\mathrm{Q}[Y=1]=q$, 
the relative frequency of the positive labels in test sample is represented by the random variable
\begin{equation}\label{eq:RelFreq}
\widehat{Y}_{n, \mathrm{Q}} \ = \ \frac{1}{n} \sum_{i=1}^n I(Y_{i, \mathrm{Q}}=1),
\end{equation}
where $I(Y_{i, \mathrm{Q}}=1) = 1$ if $Y_{i, \mathrm{Q}}=1$ and $I(Y_{i, \mathrm{Q}}=1) = 0$ otherwise.

The simulation procedures for the panels of Table~\ref{tab:pred} are intended to gauge 
the impact of using a confidence interval instead
of a prediction interval for capturing the relative frequency of positive labels in the test sample as defined
in \eqref{eq:RelFreq}.
By the law of large numbers, the difference of  $\widehat{Y}_{n, \mathrm{Q}}$ and $q$ ought to be small
for large $n$. Therefore, if there is any impact of using a confidence interval when a prediction interval
would be needed, it should rather be visible for smaller $n$. 

The algorithm devised in this paper for the construction of prediction intervals (see Section~\ref{sec:calc})
involves the simulation of binomial random variables with the prevalence estimates
as success probabilities which are independent of the test samples. This procedure, however, is likely
to exaggerate the variance of the relative frequencies of the positive labels because
the prevalence estimates and the test samples are not only not independent but even by design 
should be strongly dependent. The dependence between prevalence estimate and the test sample should
be the stronger, the more accurate the classifier underlying the estimator is. This implies that
for prevalence estimation, differences between prediction and confidence intervals should rather be
discernible for lower accuracy of the classifiers deployed.

\begin{table}[t!p]
{\small \caption{Illustration of coverage of positive class frequencies in the test
sample by means of prediction and confidence intervals.
Control parameter for all panels: $n = 50$.}\label{tab:pred}
\vspace{-1ex}
\begin{center}
 \begin{tabular}{|l||r|r|r|r|r|r|r|r|r|r|}\hline
\multicolumn{11}{|c|}{$m^+=33$, $m^-=67$, $\nu=2.5$, $q=0.2$, prediction intervals}\\ \hline
\multicolumn{1}{|c||}{}&\multicolumn{1}{c}{ACC50}&\multicolumn{1}{|c}{ACCp}&\multicolumn{1}{|c}{ACCv}&
\multicolumn{1}{|c}{MS}&\multicolumn{1}{|c}{APCC}&\multicolumn{1}{|c}{APCCv}&\multicolumn{1}{|c}{H4}&
\multicolumn{1}{|c}{H8}&\multicolumn{1}{|c}{Energy}&\multicolumn{1}{|c|}{MLboot}\\\hline
Av~freq&19.26&20.70&16.02&20.72&20.42&18.74&18.92&19.72&19.76&19.38 \\ \hline
Av~abs~dev&7.50&8.82&9.02&7.60&7.58&7.58&7.40&7.24&7.28&7.50 \\ \hline
Perc fail est&0.0&0.0&0.0&0.0&0.0&0.0&0.0&0.0&0.0&0.0 \\ \hline
Av~int~length&32.20&33.30&29.98&30.36&30.04&29.68&30.32&29.86&30.00&29.88 \\ \hline
Coverage&100.0&99.0&94.0&100.0&99.0&98.0&99.0&99.0&99.0&98.0 \\ \hline
Perc 0 or 1&1.0&2.0&6.0&1.0&2.0&2.0&4.0&1.0&3.0&1.0 \\ \hline
\multicolumn{11}{|c|}{$m^+=33$, $m^-=67$, $\nu=2.5$, $q=0.2$, confidence intervals}\\ \hline
\multicolumn{1}{|c||}{}&\multicolumn{1}{c}{ACC50}&\multicolumn{1}{|c}{ACCp}&\multicolumn{1}{|c}{ACCv}&
\multicolumn{1}{|c}{MS}&\multicolumn{1}{|c}{APCC}&\multicolumn{1}{|c}{APCCv}&\multicolumn{1}{|c}{H4}&
\multicolumn{1}{|c}{H8}&\multicolumn{1}{|c}{Energy}&\multicolumn{1}{|c|}{MLboot}\\\hline
Av~prev&20.34&21.05&17.01&20.50&20.54&20.34&20.63&20.31&20.55&20.65 \\ \hline
Av~abs~dev&6.68&7.23&7.61&6.22&6.12&6.13&6.06&5.88&6.18&6.00 \\ \hline
Perc fail est&0.0&0.0&0.0&0.0&0.0&0.0&0.0&0.0&0.0&0.0 \\ \hline
Av~int~length&28.27&28.98&26.13&25.67&25.15&24.55&25.57&25.10&25.10&24.83 \\ \hline
Coverage&97.0&97.0&89.0&98.0&97.0&96.0&98.0&95.0&97.0&98.0 \\ \hline
Perc 0 or 1&0.0&2.0&1.0&0.0&0.0&1.0&1.0&0.0&0.0&1.0 \\ \hline\hline
\multicolumn{11}{|c|}{$m^+=67$, $m^- = 33$, $\nu=1$, $q=0.2$, prediction intervals}\\ \hline
\multicolumn{1}{|c||}{}&\multicolumn{1}{c}{ACC50}&\multicolumn{1}{|c}{ACCp}&\multicolumn{1}{|c}{ACCv}&
\multicolumn{1}{|c}{MS}&\multicolumn{1}{|c}{APCC}&\multicolumn{1}{|c}{APCCv}&\multicolumn{1}{|c}{H4}&
\multicolumn{1}{|c}{H8}&\multicolumn{1}{|c}{Energy}&\multicolumn{1}{|c|}{MLboot}\\\hline
Av~freq&18.96&22.30&23.72&21.33&17.08&18.20&22.80&28.76&19.12&17.48 \\ \hline
Av~abs~dev&19.00&16.98&17.32&15.56&15.16&14.32&15.12&17.20&15.04&14.48 \\ \hline
Perc fail est&0.0&0.0&0.0&1.0&0.0&0.0&0.0&0.0&0.0&0.0 \\ \hline
Av~int~length&71.69&58.92&72.82&54.15&47.88&49.96&56.24&59.32&49.44&47.38 \\ \hline
Coverage&95.0&94.0&99.0&95.0&92.0&94.0&92.0&87.0&93.0&91.0 \\ \hline
Perc 0 or 1&43.0&26.0&19.0&17.2&32.0&24.0&20.0&9.0&24.0&24.0 \\ \hline
\multicolumn{11}{|c|}{$m^+=67$, $m^- = 33$, $\nu=1$, $q=0.2$, confidence intervals}\\ \hline
\multicolumn{1}{|c||}{}&\multicolumn{1}{c}{ACC50}&\multicolumn{1}{|c}{ACCp}&\multicolumn{1}{|c}{ACCv}&
\multicolumn{1}{|c}{MS}&\multicolumn{1}{|c}{APCC}&\multicolumn{1}{|c}{APCCv}&\multicolumn{1}{|c}{H4}&
\multicolumn{1}{|c}{H8}&\multicolumn{1}{|c}{Energy}&\multicolumn{1}{|c|}{MLboot}\\\hline
Av~prev&21.36&22.90&21.16&23.61&19.25&20.93&23.65&28.81&21.26&19.59 \\ \hline
Av~abs~dev&19.28&16.45&15.21&14.60&14.63&13.91&15.64&15.53&14.56&13.37 \\ \hline
Perc fail est&0.0&0.0&0.0&4.0&0.0&0.0&0.0&0.0&0.0&0.0 \\ \hline
Av~int~length&66.65&57.89&69.97&51.98&47.20&49.07&54.25&54.26&48.45&47.08 \\ \hline
Coverage&97.0&95.0&98.0&96.0&90.0&95.0&90.0&77.0&94.0&96.0 \\ \hline
Perc 0 or 1&36.0&24.0&22.0&15.6&27.0&22.0&19.0&7.0&22.0&16.0 \\ \hline\hline
\multicolumn{11}{|c|}{$m^+=67$, $m^- = 33$, $\nu=1$, $q=0.05$, prediction intervals}\\ \hline
\multicolumn{1}{|c||}{}&\multicolumn{1}{c}{ACC50}&\multicolumn{1}{|c}{ACCp}&\multicolumn{1}{|c}{ACCv}&
\multicolumn{1}{|c}{MS}&\multicolumn{1}{|c}{APCC}&\multicolumn{1}{|c}{APCCv}&\multicolumn{1}{|c}{H4}&
\multicolumn{1}{|c}{H8}&\multicolumn{1}{|c}{Energy}&\multicolumn{1}{|c|}{MLboot}\\\hline
AAv~freq&13.90&10.78&12.48&12.02&10.50&10.96&13.74&19.86&11.80&8.48 \\ \hline
Av~abs~dev&13.06&10.34&11.84&10.94&10.36&10.44&12.88&16.52&11.12&8.44 \\ \hline
Perc fail est&0.0&0.0&0.0&1.0&0.0&0.0&0.0&0.0&0.0&0.0 \\ \hline
Av~int~length&62.61&44.90&59.71&44.94&39.12&41.48&46.78&52.30&41.30&35.30 \\ \hline
Coverage&98.0&97.0&97.0&94.0&93.0&92.0&90.0&75.0&92.0&93.0 \\ \hline
Perc 0 or 1&41.0&42.0&42.0&36.4&47.0&42.0&39.0&13.0&38.0&47.0 \\ \hline
\multicolumn{11}{|c|}{$m^+=67$, $m^- = 33$, $\nu=1$, $q=0.05$, confidence intervals}\\ \hline
\multicolumn{1}{|c||}{}&\multicolumn{1}{c}{ACC50}&\multicolumn{1}{|c}{ACCp}&\multicolumn{1}{|c}{ACCv}&
\multicolumn{1}{|c}{MS}&\multicolumn{1}{|c}{APCC}&\multicolumn{1}{|c}{APCCv}&\multicolumn{1}{|c}{H4}&
\multicolumn{1}{|c}{H8}&\multicolumn{1}{|c}{Energy}&\multicolumn{1}{|c|}{MLboot}\\\hline
Av~prev&13.43&11.88&11.65&12.15&9.52&10.40&12.77&18.30&10.70&8.33 \\ \hline
Av~abs~dev&13.79&11.11&11.03&10.68&9.56&9.71&11.44&15.43&10.01&8.34 \\ \hline
Perc fail est&0.0&0.0&0.0&4.0&0.0&0.0&0.0&0.0&0.0&0.0 \\ \hline
Av~int~length&58.34&45.14&55.64&41.20&34.38&37.22&43.41&48.10&37.30&32.95 \\ \hline
Coverage&98.0&93.0&96.0&93.0&89.0&91.0&83.0&62.0&94.0&92.0 \\ \hline
Perc 0 or 1&52.0&40.0&42.0&31.2&42.0&36.0&34.0&17.0&37.0&44.0 \\ \hline
\end{tabular} 
\end{center} }
\end{table}

Table~\ref{tab:pred} shows a number of simulation results, all for test sample size $n=50$, i.e.\ for
small size of the test sample:
\begin{itemize}
\item Top two panels: Simulation of a `benign' situation, with not too much 
difference of positive class prevalences (33\% vs.\ 20\%)
in training and test population distributions, and high power of the score underlying the classifiers and distance 
minimisation approaches. Results suggest `overshooting' by the binomial prediction interval approach, i.e.\ 
intervals are so long that coverage is much higher than requested, even reaching 100\%. The confidence intervals
clearly show sufficient coverage of the true realised percentages of positive labels 
for all estimation methods. Interval lengths are quite uniform, with only
the straight ACC methods ACC50 and ACCp showing distinctly longer intervals. Also in terms of 
average absolute deviation from the true positive class prevalence, the performance is rather uniform. However,
it is interesting to see that ACCv which has been designed for minimising confidence interval length 
among the ACC estimators shows the distinctly worst performance with regard to average absolute deviation.
\item Central two panels: Simulation of a rather adverse situation, with very different (67\% vs. 20\%)
positive class prevalences in training and test population distributions and low power of the 
score underlying the classifiers and distance 
minimisation approaches. There is still overshooting
by the binomial prediction interval approach for all methods but H8. For all methods but H8
sufficient coverage by the confidence intervals is still clearly achieved. H8 coverage of relative positive class
frequency is significantly too low with the confidence intervals but still sufficient with the prediction 
intervals. However, H8 also displays heavy bias of the average relative frequency of positive labels, possibly
a consequence of the combined difficulties of there being 8 bins for only 50 points (test sample size)
and little difference between the densities of the score conditional on the two classes. In terms of
interval length performance MLboot is best, closely followed by APCC and Energy. But even for these methods,
confidence interval lengths of more then 47\% suggest that the estimation task is rather hopeless.
\item Bottom two panels: Similar picture to the central panels, but even more adverse with a small test 
sample prevalence of 5\%. Results similar, but much higher proportions of 0\% estimates for all methods. H8 now 
has insufficient coverage with both prediction and confidence
intervals, and also H4 coverage with the confidence intervals is insufficient. Note the strong estimation bias 
suggested by all average frequency estimates, presumably caused by the clipping of negative estimates (i.e.\ 
replacing such estimates by zero). Among all these bad estimators, MLboot is clearly best in terms of 
bias, average absolute deviation and interval lengths.
\item General conclusion: For all methods from Section~\ref{sec:approaches} but the Hellinger methods, it suffices
to construct confidence intervals. No need to apply special prediction interval techniques. 
\item Performance in terms of interval length (with sufficient coverage in all circumstances): MLboot best, followed by
APCC and Energy.
\end{itemize}

\begin{table}[t!p]
{\small \caption{Illustration of difference between binomial approach and 
approach by \citet{daughton2019constructing} to prediction intervals for positive class prevalences.
Control parameters for both panels: $n = 50$, $\nu=2.5$, $q=0.2$.
Columns `ACC50' and `ACCp' show results for confidence intervals with methods as explained in Section~\ref{sec:approaches}.
All other columns show prediction intervals and average relative frequencies of positive labels.
In all columns, coverage refers to containing the realised relative frequency of positive labels
in the test sample.
`DnPACC50' and `DnPACCp' are determined according to the `error-corrected bootstrapping' as 
proposed by \citeauthor{daughton2019constructing}.}
\label{tab:DnP}
%\vspace{1ex}
\begin{center}
  \begin{tabular}{|l||r|r|r|r|r|r|}\hline
  \multicolumn{7}{|c|}{$m^+=33$, $m^-=67$}\\ \hline
  \multicolumn{1}{|c||}{}&\multicolumn{1}{c}{ACC50}&\multicolumn{1}{|c}{predACC50}&\multicolumn{1}{|c}{ACCp}&
\multicolumn{1}{|c}{predACCp}&\multicolumn{1}{|c}{DnPACC50}&\multicolumn{1}{|c|}{DnPACCp}\\\hline
Av~prev or freq&19.97&20.22&19.49&19.78&24.32&25.40 \\ \hline
Av~abs~dev&6.19&7.70&6.56&8.54&5.68&6.80 \\ \hline
Perc fail est&0.0&0.0&0.0&0.0&0.0&0.0 \\ \hline
Av~int~length&27.91&32.84&29.28&33.90&21.68&21.78 \\ \hline
Coverage&100.0&100.0&99.0&100.0&95.0&93.0 \\ \hline
Perc 0 or 1&0.0&0.0&1.0&2.0&0.0&0.0 \\ \hline\hline
 \multicolumn{7}{|c|}{$m^+=67$, $m^-=33$}\\ \hline
\multicolumn{1}{|c||}{}&\multicolumn{1}{c}{ACC50}&\multicolumn{1}{|c}{predACC50}&\multicolumn{1}{|c}{ACCp}&
\multicolumn{1}{|c}{predACCp}&\multicolumn{1}{|c}{DnPACC50}&\multicolumn{1}{|c|}{DnPACCp}\\\hline
Av~prev&19.69&19.38&19.58&20.28&38.06&39.26 \\ \hline
Av~abs~dev&9.91&9.82&9.03&10.32&18.18&19.26 \\ \hline
Perc fail est&0.0&0.0&0.0&0.0&0.0&0.0 \\ \hline
Av~int~length&32.41&36.34&30.49&34.72&26.28&26.32 \\ \hline
Coverage&98.0&99.0&95.0&98.0&22.0&14.0 \\ \hline
Perc 0 or 1&12.0&12.0&3.0&5.0&0.0&0.0 \\ \hline
\end{tabular}
\end{center} }
\end{table}

\citet{daughton2019constructing} proposed `Error Adjusted Bootstrapping' as an approach
to constructing ``confidence intervals'' (prediction intervals, as a matter of fact)  
for prevalences and showed by example that its
performance in terms of coverage was sufficient. However, theoretical analysis of `Error Adjusted Bootstrapping'
presented in Appendix~\ref{sec:appB} suggests that this approach is not appropriate
for constructing prediction intervals in the presence of prior probability shift. Indeed,
Table~\ref{tab:DnP} demonstrates that  `Error Adjusted Bootstrapping' intervals based 
on the classifiers ACC50 and ACCp (see Section~\ref{sec:approaches}) achieve sufficient coverage
if the difference between the training and test sample prevalences is moderate (33\% vs.\ 20\%) but
breaks down if the difference is large (67\% vs.\ 20\%).

\begin{table}[t!p]
{\small \caption{Illustration of length of confidence intervals for different degrees
of accuracy (or discriminatory power) of the score or classifier underlying the
prevalence estimation methods.}\label{tab:power}
%\vspace{1ex}
\begin{center}
 \begin{tabular}{|l||r|r|r|r|r|r|r|r|r|r|}\hline
\multicolumn{11}{|c|}{$n=500$, $m^+=\infty$, $m^-=\infty$, $\nu=2.5$, $p=0.5$, $q=0.2$}\\ \hline
\multicolumn{1}{|c||}{}&\multicolumn{1}{c}{ACC50}&\multicolumn{1}{|c}{ACCp}&\multicolumn{1}{|c}{ACCv}&
\multicolumn{1}{|c}{MS}&\multicolumn{1}{|c}{APCC}&\multicolumn{1}{|c}{APCCv}&\multicolumn{1}{|c}{H4}&
\multicolumn{1}{|c}{H8}&\multicolumn{1}{|c}{Energy}&\multicolumn{1}{|c|}{MLinf}\\\hline
Av~prev&20.28&20.28&19.68&20.34&20.35&20.27&20.27&20.33&20.34&20.35 \\ \hline
Av~abs~dev&1.97&1.97&1.90&1.79&1.79&1.72&1.81&1.76&1.80&1.72 \\ \hline
Perc fail est&0.0&0.0&0.0&0.0&0.0&0.0&0.0&0.0&0.0&0.0 \\ \hline
Av~int~length&8.47&8.47&8.04&7.74&7.71&7.50&7.54&7.42&7.72&7.35 \\ \hline
Coverage&91.0&91.0&93.0&90.0&89.0&92.0&89.0&89.0&89.0&91.0 \\ \hline
Perc 0 or 1&0.0&0.0&0.0&0.0&0.0&0.0&0.0&0.0&0.0&0.0 \\ \hline
\multicolumn{11}{|c|}{$n=500$, $m^+=\infty$, $m^-=\infty$, $\nu=1$, $p=0.5$, $q=0.2$}\\ \hline
\multicolumn{1}{|c||}{}&\multicolumn{1}{c}{ACC50}&\multicolumn{1}{|c}{ACCp}&\multicolumn{1}{|c}{ACCv}&
\multicolumn{1}{|c}{MS}&\multicolumn{1}{|c}{APCC}&\multicolumn{1}{|c}{APCCv}&\multicolumn{1}{|c}{H4}&
\multicolumn{1}{|c}{H8}&\multicolumn{1}{|c}{Energy}&\multicolumn{1}{|c|}{MLinf}\\\hline
Av~prev&20.71&20.71&18.60&20.74&20.53&20.00&20.47&20.47&20.66&20.41 \\ \hline
Av~abs~dev&4.68&4.68&4.21&3.64&3.37&3.23&3.65&3.30&3.39&3.19 \\ \hline
Perc fail est&0.0&0.0&0.0&0.0&0.0&0.0&0.0&0.0&0.0&0.0 \\ \hline
Av~int~length&19.08&19.08&18.47&16.70&15.73&15.15&16.14&15.50&15.91&15.05 \\ \hline
Coverage&92.0&92.0&93.0&94.0&94.0&94.0&96.0&95.0&95.0&94.0 \\ \hline
Perc 0 or 1&0.0&0.0&1.0&0.0&0.0&0.0&0.0&0.0&0.0&0.0 \\ \hline\hline
\multicolumn{11}{|c|}{$n=50$, $m^+=\infty$, $m^-=\infty$, $\nu=2.5$, $p=0.33$, $q=0.05$}\\ \hline
\multicolumn{1}{|c||}{}&\multicolumn{1}{c}{ACC50}&\multicolumn{1}{|c}{ACCp}&\multicolumn{1}{|c}{ACCv}&
\multicolumn{1}{|c}{MS}&\multicolumn{1}{|c}{APCC}&\multicolumn{1}{|c}{APCCv}&\multicolumn{1}{|c}{H4}&
\multicolumn{1}{|c}{H8}&\multicolumn{1}{|c}{Energy}&\multicolumn{1}{|c|}{MLinf}\\\hline
Av~prev&4.96&5.49&2.75&4.99&5.14&4.60&5.07&4.56&5.15&5.09 \\ \hline
Av~abs~dev&3.63&4.03&3.26&3.32&3.47&2.92&3.33&2.93&3.54&3.06 \\ \hline
Perc fail est&0.0&0.0&0.0&0.0&0.0&0.0&0.0&0.0&0.0&0.0 \\ \hline
Av~int~length&16.75&17.98&14.44&12.65&12.73&10.88&12.95&12.03&12.84&16.96 \\ \hline
Coverage&95.0&98.0&97.0&90.0&86.0&85.0&88.5&94.4&88.0&97.0 \\ \hline
Perc 0 or 1&24.0&24.0&27.0&16.0&18.0&11.0&18.0&10.0&18.0&13.0 \\ \hline
\multicolumn{11}{|c|}{$n=50$, $m^+=\infty$, $m^-=\infty$, $\nu=1$, $p=0.33$, $q=0.05$}\\ \hline
\multicolumn{1}{|c||}{}&\multicolumn{1}{c}{ACC50}&\multicolumn{1}{|c}{ACCp}&\multicolumn{1}{|c}{ACCv}&
\multicolumn{1}{|c}{MS}&\multicolumn{1}{|c}{APCC}&\multicolumn{1}{|c}{APCCv}&\multicolumn{1}{|c}{H4}&
\multicolumn{1}{|c}{H8}&\multicolumn{1}{|c}{Energy}&\multicolumn{1}{|c|}{MLinf}\\\hline
Av~prev&8.37&9.56&2.77&8.07&7.74&6.56&7.86&7.37&7.93&7.26 \\ \hline
Av~abs~dev&8.31&9.28&5.70&7.78&7.64&7.00&7.73&7.47&7.68&7.50 \\ \hline
Perc fail est&0.0&0.0&0.0&0.0&0.0&0.0&0.0&0.0&0.0&0.0 \\ \hline
Av~int~length&38.27&35.30&29.85&27.52&25.46&24.98&27.63&25.80&26.30&47.01 \\ \hline
Coverage&96.0&92.0&90.0&89.0&86.0&86.0&89.0&86.3&89.0&97.0 \\ \hline
Perc 0 or 1&38.0&44.0&65.0&32.0&41.0&48.0&41.0&45.0&43.0&48.0 \\ \hline\hline
\multicolumn{11}{|c|}{$n=50$, $m^+=33$, $m^-=67$, $\nu=2.5$, $q=0.05$}\\ \hline
\multicolumn{1}{|c||}{}&\multicolumn{1}{c}{ACC50}&\multicolumn{1}{|c}{ACCp}&\multicolumn{1}{|c}{ACCv}&
\multicolumn{1}{|c}{MS}&\multicolumn{1}{|c}{APCC}&\multicolumn{1}{|c}{APCCv}&\multicolumn{1}{|c}{H4}&
\multicolumn{1}{|c}{H8}&\multicolumn{1}{|c}{Energy}&\multicolumn{1}{|c|}{MLboot}\\\hline
Av~prev&5.72&6.91&3.62&5.67&5.88&5.72&5.53&5.65&5.97&5.59 \\ \hline
Av~abs~dev&4.46&5.10&3.54&4.11&4.11&3.97&4.00&3.95&4.18&3.55 \\ \hline
Perc fail est&0.0&0.0&0.0&0.0&0.0&0.0&0.0&0.0&0.0&0.0 \\ \hline
Av~int~length&16.62&18.68&11.35&14.96&14.92&13.90&14.22&13.44&15.09&13.95 \\ \hline
Coverage&89.0&87.0&91.0&88.0&86.0&87.0&84.8&82.0&85.0&85.0 \\ \hline
Perc 0 or 1&24.0&22.0&24.0&18.0&17.0&17.0&18.0&15.0&17.0&14.0 \\ \hline
\multicolumn{11}{|c|}{$n=50$, $m^+=33$, $m^-=67$, $\nu=1$, $q=0.05$}\\ \hline
\multicolumn{1}{|c||}{}&\multicolumn{1}{c}{ACC50}&\multicolumn{1}{|c}{ACCp}&\multicolumn{1}{|c}{ACCv}&
\multicolumn{1}{|c}{MS}&\multicolumn{1}{|c}{APCC}&\multicolumn{1}{|c}{APCCv}&\multicolumn{1}{|c}{H4}&
\multicolumn{1}{|c}{H8}&\multicolumn{1}{|c}{Energy}&\multicolumn{1}{|c|}{MLboot}\\\hline
Av~prev&10.49&11.59&6.54&10.43&8.50&8.38&10.96&12.12&9.50&8.39 \\ \hline
Av~abs~dev&10.47&10.94&6.58&9.55&8.30&7.88&10.00&10.43&8.83&7.72 \\ \hline
Perc fail est&0.0&0.0&0.0&4.0&0.0&0.0&0.0&0.0&0.0&0.0 \\ \hline
Av~int~length&52.48&46.04&40.00&36.00&32.37&31.70&39.38&39.11&32.85&32.96 \\ \hline
Coverage&97.0&95.0&96.0&94.0&91.0&90.0&91.0&92.0&92.0&92.0 \\ \hline
Perc 0 or 1&43.0&37.0&43.0&32.3&41.0&38.0&36.0&28.0&40.0&38.0 \\ \hline
\end{tabular} 
\end{center} }
\end{table}

\subsection{Does higher accuracy help for shorter confidence intervals?}
\label{sec:shorter}

As mentioned in Section~\ref{se:intro}, views in the literature differ on whether or not the performance
of prevalence estimators is impacted by the discriminatory power of the score underlying the estimation
method.
Table~\ref{tab:power} shows a number of simulation results, for a variety of sets of circumstances, both
benign and adverse. Results for high and low power are juxtaposed:
\begin{itemize}
\item Top two panels: Simulation of a `benign' situation, with moderate
difference of positive class prevalences (50\% vs.\ 20\%)
in training and test population distributions, no estimation uncertainty on the training sample and a
rather large test sample with $n=500$. Results for all estimation methods suggest that the lengths of
the confidence intervals are strongly dependent upon the discriminatory power of the score which is the basic
building block of all the methods. Coverage is accurate for the high power situation whereas there
is even slight overshooting of coverage in the low power situation.
\item Central two panels: Simulation of a less benign situation, with small test sample size and
low true positive class prevalence in the test sample but still without uncertainty on the training
sample. 
There is nonetheless again evidence for the strong dependence of the lengths of
the confidence intervals upon the discriminatory power of the score.
For all estimation methods, low power leads to strong bias of the prevalence estimates.
The percentage of zero estimates jumps between the 3rd and the 4th panel. Hence, decrease of power of
the score entails much higher rates of zero estimates. For the maximum likelihood
method, the interval length results in both panels show that constructing confidence intervals 
based on the central limit theorem for
maximum likelihood estimators may become unstable for small test sample size and small positive
class prevalence. 
\item Bottom two panels: Simulation of an adverse situation, with small test and training sample sizes and
low true positive class prevalence in the test sample. Results show qualitatively very much the same picture
as in the central panels. The impact of estimation uncertainty in the training sample which marks the
difference to the situation for the central panels, however, is moderate for high power of the score but
dramatic for low power of the score. Again there is a jump of the rate of zero estimates between
the two panels differentiated by different levels of discriminatory power.  For the Hellinger methods,
results of the high power panel suggest a performance issue\footnote{%
This observation is not confirmed by a repetition of the calculations for Panel~5 with
deployment of R-boot.ci method `bca` instead of `perc'. However, the `better' results
are accompanied by a high rate of failures of the confidence interval construction.} 
with respect to the coverage rate. In 
contrast to MLinf, MLboot (using only bootstrapping for constructing the confidence intervals)
performs well, even with relatively low bias for the prevalence estimate in the low power case.
\item General conclusion:   The results displayed in Table~\ref{tab:power} suggest
that there should be a clear benefit in terms of shorter confidence intervals 
when high power scores and classifiers are deployed for prevalence estimation. 
In addition, the results illustrate the statement on the asymptotic variance of ratio estimators 
like ACC50, ACCp, APCv, APCC and APCCv in Corollary~11 of \citet{Vaz&Izbicki&Stern2019}.
\item Performance in terms of interval length (with sufficient coverage in all circumstances): Both
APCC estimation methods show good and stable performance when compared to all other methods. Energy
and MLboot follow closely. The Hellinger methods also produce short confidence lengths but 
may have insufficient coverage. 
\end{itemize}

\subsection{Do approaches to confidence intervals without Monte Carlo simulations work?}
\label{sec:NoSim}

For the prevalence estimation methods ACC50, ACCp, ACCv, MS, and MLinf, confidence intervals can be constructed without
bootstrapping and, therefore, much less numerical effort.
For ACC50, ACCp, ACCv, and MS,  conservative binomial intervals by means of the `exact' method of R-function binconf
can be deployed  \citep[][Section~6.2.2]{MeekerEtAl}.
For the maximum likelihood approach, an asymptotically most efficient normal approximation with variance expressed in 
terms of the Fisher information can be used \citep[Theorem 10.1.12,][]{Casella&Berger}. This approach
is denoted by `MLinf' in order to distinguish it from `MLboot`, maximum likelihood estimation combined with
bootstrapping for the confidence intervals.

However, it can be shown by examples that these non-simulation approaches fail in the sense of producing 
insufficient coverage rates if training sample sizes are finite, i.e.\ if parameters like true positive 
and false positive rates needed for the estimators have to be estimated (e.g.\ by means of 
regression) before being plugged in.
Table~\ref{tab:NoSim} with panels juxtaposing results for infinite sample and finite sample sizes of the training sample, 
provides such an example.

The estimation problem whose results are shown in Table~\ref{tab:NoSim} is pretty well-posed, with 
a large test sample, a high power score underlying the estimation methods and moderate difference
between training and test sample positive class prevalences. Panel~1 shows that without 
estimation uncertainty on the training sample (infinite sample size) the non-simulation approaches
produce confidence intervals with sufficient coverage. In contrast, Panel~2 demonstrates that
for all five methods coverage breaks down when estimation uncertainty is introduced into 
the training sample (finite sample size). According to Panel~3, this issue can be remediated 
by deploying bootstrapping for the construction of the confidence intervals.

\begin{table}[t!p]
{\small \caption{Illustration of failure of non-simulation approaches to 
confidence intervals when training sample is finite.
Control parameters for all panels: $p=0.33$, $\nu=2.5$, $q=0.2$, $n=500$.}
\label{tab:NoSim}
%\vspace{1ex}
\begin{center}
  \begin{tabular}{|l||r|r|r|r|r|r|}\hline
  \multicolumn{6}{|c|}{No bootstrap, $m^+=\infty$, $m^-=\infty$}\\ \hline
  \multicolumn{1}{|c||}{}&\multicolumn{1}{c}{ACC50}&\multicolumn{1}{|c}{ACCp}&
\multicolumn{1}{|c}{ACCv}&\multicolumn{1}{|c}{MS}&\multicolumn{1}{|c|}{MLinf}\\\hline
Av~prev&20.15&20.28&19.40&20.14&20.27 \\ \hline
Av~abs~dev&2.25&2.24&2.18&2.22&2.02 \\ \hline
Perc fail est&0.0&0.0&0.0&0.0&0.0 \\ \hline
Av~int~length&8.11&8.46&8.02&8.17&7.33 \\ \hline
Coverage&92.0&86.0&88.0&87.0&92.0 \\ \hline
Perc 0 or 1&0.0&0.0&0.0&0.0&0.0 \\ \hline
 \multicolumn{6}{|c|}{No bootstrap, $m^+=33$, $m^-=67$}\\ \hline
  \multicolumn{1}{|c||}{}&\multicolumn{1}{c}{ACC50}&\multicolumn{1}{|c}{ACCp}&
\multicolumn{1}{|c}{ACCv}&\multicolumn{1}{|c}{MS}&\multicolumn{1}{|c|}{MLinf}\\\hline
Av~prev&19.95&19.48&19.01&19.76&20.06 \\ \hline
Av~abs~dev&3.23&3.88&2.98&3.35&2.57 \\ \hline
Perc fail est&0.0&0.0&0.0&0.0&0.0 \\ \hline
Av~int~length&8.13&8.47&7.60&8.16&7.22 \\ \hline
Coverage&69.0&64.0&66.0&66.0&75.0 \\ \hline
Perc 0 or 1&0.0&0.0&0.0&0.0&0.0 \\ \hline
 \multicolumn{6}{|c|}{Bootstrap, $m^+=33$, $m^-=67$}\\ \hline
  \multicolumn{1}{|c||}{}&\multicolumn{1}{c}{ACC50}&\multicolumn{1}{|c}{ACCp}&
\multicolumn{1}{|c}{ACCv}&\multicolumn{1}{|c}{MS}&\multicolumn{1}{|c|}{MLboot}\\\hline
Av~prev&19.79&19.94&18.63&20.18&20.34 \\ \hline
Av~abs~dev&3.11&3.38&3.17&2.79&2.67 \\ \hline
Perc fail est&0.0&0.0&0.0&0.0&0.0 \\ \hline
Av~int~length&15.13&16.96&14.18&12.94&12.07 \\ \hline
Coverage&95.0&93.0&91.0&91.0&92.0 \\ \hline
Perc 0 or 1&0.0&0.0&0.0&0.0&0.0 \\ \hline
\end{tabular}
\end{center} }
\end{table}

%%%
% New section    
%%%

\section{Conclusions}
\label{se:concl}

The simulation study whose results are reported in this paper has been intended to shed some 
light on certain questions from the literature regarding the construction of confidence or prediction intervals
for the prevalence of positive labels in binary quantification problems. In particular, 
the results of the study should help to provide answers to the questions of
\begin{itemize}
 \item whether estimation techniques for confidence intervals are appropriate if in practice
 most of the time prediction intervals are needed, and
 \item whether the discriminatory power of the soft classifier or score at the basis of a prevalence estimation method
 matters when it comes to minimizing the confidence interval for an estimate.
 \end{itemize} 
The answers suggested by the results of the simulation study are subject to a number of qualifications. Most
prominent among the qualifications are
\begin{itemize}
\item the fact that the findings of the paper apply only for problems where it is clear that training and test sample
are related by prior probability shift, and
\item the general observation that the scope of a simulation study necessarily is rather restricted and therefore
findings of such studies can be suggestive and illustrative at best. 
\end{itemize}
Hence the findings from the study do not allow firm or general conclusions. As a consequence, the answers
to the questions suggested by the simulation study have to be ingested with caution:
\begin{itemize}
\item For not too small test sample sizes like 50 or more, there is no need to deploy special
techniques for prediction intervals.
\item It is worthwhile to base prevalence estimation on powerful classifiers or scores because this
way the lengths of the confidence intervals can be much reduced. The use of less accurate classifiers
may entail confidence intervals so long that the estimates have to be considered worthless.
\end{itemize}
In most of the experiments performed as part of the simulation study, the maximum likelihood approach (method MLboot) to
the estimation of the positive class prevalence turned out to deliver on average the shortest confidence intervals.
As shown in Appendix~\ref{sec:KL}, application of the maximum likelihood approach requires that in a previous step
the density ratio or the posterior class probabilities are estimated on the training samples. To achieve
this with sufficient precision is a notoriously hard problem. Note, however, the promising recent
progress made on this issue \citep{kull2017betacalibration}. Not much worse and in a few cases even superior
was the performance of APCC (Adjusted Probabilistic Classify \& Count). In contrast the performance of the
Energy distance and Hellinger distance estimation methods was not outstanding and, in the case of the latter
methods, even insufficient in the sense of not guaranteeing the required coverage rates of the confidence intervals.

Recent research by \citet{Maletzke&etAl.Quantification} singled out prevalence estimation methods based
on minimising distances related to the Earth Mover's distance as very well and robustly performing.
Earlier research by \citet{hofer2015adapting} already found that prevalence estimation by minimising 
the Earth Mover's distance worked well in the presence of general data set shift (`local drift').
Hence it might be worthwhile to compare the performance of such estimators with respect to the
length of confidence intervals to the performance of other estimators like the ones considered in this paper.

%%%
% References
%%%

%\bibliographystyle{plainnat}
%\bibliography{Literature}

\addcontentsline{toc}{section}{References}

%%%
% Appendices
%%%
\appendix
\section{Appendix: Particulars for the implementation of the simulation study}
\label{sec:app}
%\addcontentsline{toc}{section}{Particulars for the case of an infinite training sample}

This appendix presents the mathematical details needed for coding the 
prevalence estimation methods listed in Section~\ref{sec:approaches}. In particular,
the case of infinite training samples (i.e.\ where the training sample is actually
the training population and the parameters of the model are exactly known) is covered.

\subsection{Adjusted Classify \& Count (ACC) and related prevalence estimators}
\label{sec:ACCetRel}

ACC and APCC as mentioned in Section~\ref{sec:approaches} are special
cases of the `ratio estimator' of \citet{Vaz&Izbicki&Stern2019}. 
From an even more
general perspective, they are instances of estimation by the `Method of Moments'
\citep[][Section~2.4.1 and the references therein]{fruhwirth2006finite}. 
By Theorem~6 of \citet{Vaz&Izbicki&Stern2019},
ratio estimators are Fisher consistent for
estimating the positive class prevalence of the test population under prior probability shift.

\textbf{Adjusted Classify \& Count (ACC).} In the setting of Section~\ref{sec:setting}, 
denote the feature space (i.e.\ the range of values which the feature variable $X$ can take) by $\mathcal{X}$.
Let $g: \mathcal{X} \to \{-1, 1\}$ be a \emph{crisp classifier} in the sense that if for an instance it holds that 
$g(X) =1$,  a positive class label is predicted, and
if $g(X) =-1$ a negative class label is predicted.
With the notation introduced in 
Section~\ref{sec:setting}, the \emph{ACC estimator} $\widehat{\mathrm{Q}}_g[Y=1]$ \emph{based on the classifier $g$}  of 
the test population positive class prevalence is given by
\begin{equation}\label{eq:ACC}
\widehat{\mathrm{Q}}_g[Y=1] \ = \ \frac{\mathrm{Q}[g(X) = 1]-\mathrm{P}[g(X)=1\,|\,Y=-1]}
    {\mathrm{P}[g(X)=1\,|\,Y=1]-\mathrm{P}[g(X)=1\,|\,Y=-1]}.
\end{equation}
Recall that 
\begin{itemize}
\item $\mathrm{Q}[g(X) = 1]$ is the proportion of instances in the test population 
whose labels are predicted positive by the classifier $g$.
\item $\mathrm{P}[g(X)=1\,|\,Y=-1]$ is the \emph{false positive rate (FPR)} associated with the classifier $g$. The FPR
equals $100\% - \text{true negative rate}$ and, therefore, also $100\% - \text{specificity}$ of the classifier $g$.
\item $\mathrm{P}[g(X)=1\,|\,Y=1]$ is the \emph{true positive rate (TPR)} associated with the classifier $g$. The TPR is also 
called `recall' or `sensitivity' of $g$.
\end{itemize}
Of course, the ACC estimator of \eqref{eq:ACC} is defined only if $\mathrm{P}[g(X)=1\,|\,Y=1] \not=\mathrm{P}[g(X)=1\,|\,Y=-1]$, i.e.\
if $g$ is not completely inaccurate.
\citet[][Section~6.2]{Gonzalez:2017:RQL:3145473.3117807} gave some background information on the history of ACC estimators.

When a threshold $t\in \mathbb{R}$ is fixed, the soft classifier $s: \mathcal{X}\to \mathbb{R}$ gives rise to 
a crisp classifier $g^{(s)}_t: \mathcal{X}\to \{-1, 1\}$, defined by
\begin{equation}\label{eq:soft}
g^{(s)}_t(x) \ = \ \begin{cases} 
    -1, & \text{if}\ s(x) < t,\\
    1, & \text{if}\ s(x) \ge t.
    \end{cases}
\end{equation}
The classifiers $p_t(x)$ with 
\begin{equation}\label{eq:BayesClass}
p_t(x) \ = \ \begin{cases} 
    -1, & \text{if}\ \mathrm{P}[Y=1\,|\,X](x) < t,\\
    1, & \text{if}\ \mathrm{P}[Y=1\,|\,X](x) \ge t.
    \end{cases}
\end{equation}
are \emph{Bayes classifiers} which minimise cost-sensitive Bayes errors, see for instance 
\citet[][Section~2.1]{tasche2017fisher}.
Thresholds of special interest are 
\begin{itemize}
\item $t=1/2$ for maximum accuracy (i.e.\ minimum classification error) which leads to the estimator ACC50
listed in Section~\ref{sec:approaches}, and
\item $t=\mathrm{P}[Y=1]$ 
for maximising the denominator of the right-hand side of \eqref{eq:ACC} which leads to the estimator ACCp
listed in Section~\ref{sec:approaches}.
\end{itemize}
For the simulation procedures run for this paper, a sample version of $\mathrm{Q}[p_t(X) = 1]$ has
been used:
\begin{equation}\label{eq:ACC.emp}
\mathrm{Q}[p_t(X) = 1] \  \approx \ \frac{1}{n} \sum_{i=1}^n I(p_t(x_{i, \mathrm{Q}})=1),
\end{equation}
where  $x_{1, \mathrm{Q}}, \ldots, x_{n, \mathrm{Q}}$ denotes a sample 
generated under the test population distribution $\mathrm{Q}(X)$.

To deal with the case where in the setting of Section~\ref{sec:model} 
with the double binormal model the training sample is infinite, 
the following formulae have been coded for 
the right-hand side of \eqref{eq:ACC} with $g(X) = p_t(X)$ and \eqref{eq:ACC.emp} 
(with parameters $a, b$ as in \eqref{eq:trainCondProb}):
\begin{align}
\frac{1}{n} \sum_{i=1}^n I(p_t(x_{i, \mathrm{Q}})=1) &\ =\ \frac{1}{n} \sum_{i=1}^n I\left(x_{i, \mathrm{Q}} \ge 
\frac{\log(1/t -1) -b}a\right),\notag\\
\mathrm{P}[p_t(X)=1\,|\,Y=-1] &\ =\ 1 - \Phi_{\mu, \sigma}\left(\frac{\log(1/t -1) -b}a\right),\label{eq:ACC.here}\\
\mathrm{P}[p_t(X)=1\,|\,Y=1] &\ =\ 1 - \Phi_{\nu, \sigma}\left(\frac{\log(1/t -1) -b}a\right).\notag
\end{align}

\textbf{Adjusted Probabilistic Classify \& Count (APCC).}
APCC -- called \emph{scaled probability average} by \citet{bella2010quantification} -- generalises \eqref{eq:ACC} by 
replacing the indicator variable $I(g(X) = 1)$ with a real-valued random variable $h(X)$. If $h$ only takes
values in the unit interval $[0,1]$ the variable $h(X)$ is a 
\emph{randomized decision classifier (RDC)} which may be interpreted
as the probability with which the positive label should be assigned. Eq.~\eqref{eq:ACC} modified for APCC reads:
\begin{equation}\label{eq:APCC}
\widehat{\mathrm{Q}}_h[Y=1] \ = \ \frac{\mathrm{E}_\mathrm{Q}[h(X)]-\mathrm{E}_\mathrm{P}[h(X)\,|\,Y=-1]}
    {\mathrm{E}_\mathrm{P}[h(X)\,|\,Y=1]-\mathrm{E}_\mathrm{P}[h(X)\,|\,Y=-1]}.
\end{equation}
\citet{bella2010quantification} suggested the choice $h(X) = \mathrm{P}[Y=1\,|\,X]$.

For the simulation procedures run for this paper, a sample version of 
$\mathrm{E}_\mathrm{Q}\bigl[\mathrm{P}[Y=1\,|\,X]\bigr]$ has been used:
\begin{equation}\label{eq:APCC.emp}
\mathrm{E}_\mathrm{Q}\bigl[\mathrm{P}[Y=1\,|\,X]\bigr] \  \approx \ 
\frac{1}{n} \sum_{i=1}^n \mathrm{P}[Y=1\,|\,X](x_{i, \mathrm{Q}}),
\end{equation}
where  $x_{1, \mathrm{Q}}, \ldots, x_{n, \mathrm{Q}}$ denotes a sample 
generated under the test population distribution $\mathrm{Q}(X)$.

To deal with the case where in the setting of Section~\ref{sec:model} 
with the double binormal model the training sample is infinite, 
the following formulae have been coded for 
the right-hand side of \eqref{eq:APCC} with $h(X) = \mathrm{P}[Y=1\,|\,X]$ and \eqref{eq:APCC.emp} 
(with parameters $a, b$ as in \eqref{eq:trainCondProb}):
\begin{align}
\frac{1}{n} \sum_{i=1}^n \mathrm{P}[Y=1\,|\,X](x_{i, \mathrm{Q}}) &\ =\ 
\frac{1}{n} \sum_{i=1}^n \frac{1}{1 + \exp(a\,x_{i, \mathrm{Q}} + b)},\notag\\
\mathrm{E}_\mathrm{P}\bigl[\mathrm{P}[Y=1\,|\,X]\,|\,Y=-1]\bigr] &\ =\ 1 - \int_{-\infty}^\infty \frac{\varphi_{\mu, \sigma}
(x_{i, \mathrm{Q}})}{1 + \exp(a\,x_{i, \mathrm{Q}} + b)}\, d\,x, \label{eq:APCC.here}\\
\mathrm{E}_\mathrm{P}\bigl[\mathrm{P}[Y=1\,|\,X]\,|\,Y=1\bigr] &\ =\ 1 - \int_{-\infty}^\infty \frac{\varphi_{\nu, \sigma}
(x_{i, \mathrm{Q}})}{1 + \exp(a\,x_{i, \mathrm{Q}} + b)}\, d\,x.\notag
\end{align}

\textbf{Median sweep (MS).} \citet{forman2008quantifying} proposed to stabilise the
prevalence estimates from ACC based on a soft classifier $s$ via \eqref{eq:soft}, by taking
the median of all ACC estimates based on $g^{(s)}_t$ for all thresholds $t$ such that 
the denominator $\mathrm{P}[g^{(s)}_t=1\,|\,Y=1]-\mathrm{P}[g^{(s)}_t=1\,|\,Y=-1]$ of the right-hand 
side of \eqref{eq:ACC} exceeds 25\%. For the purpose of this paper, the base soft classifier
is $\mathrm{P}[Y=1\,|\,X]$ in connection with \eqref{eq:BayesClass}, and the set of
possible thresholds $t$ is restricted to $\{0.05,\,0.1,\,0.15, \ldots, 0.9, \, 0.95\}$.

\textbf{Tuning ACC for ACCv.} Observe that a main factor impacting the length of a confidence interval for a parameter
is the standard deviation of the underlying estimator. This suggests the following approach to choosing a good threshold 
$t^\ast$ for the classifier $p_t(X)$ in \eqref{eq:BayesClass}: 
\begin{equation}\label{eq:ACC.tuned} 
t^\ast \ = \ \arg \min\limits_{0 < t < 1} \frac{\mathrm{var}_\mathrm{Q}[I(p_t(X)=1)]}
            {\bigl(\mathrm{P}[p_t(X)=1\,|\,Y=1]-\mathrm{P}[p_t(X)=1\,|\,Y=-1]\bigr)^2}.   
\end{equation}
The test population distribution $\mathrm{Q}$ appears in the numerator of \eqref{eq:ACC.tuned}
because the confidence interval is calculated for a sample generated from $\mathrm{Q}$.
The training population distribution $\mathrm{P}$ is used in the denominator of \eqref{eq:ACC.tuned} because 
the confidence interval is scaled by the denominator of \eqref{eq:ACC}.
See \eqref{eq:ACC.here} for the formulae used for the calculations of this paper for \eqref{eq:ACC.tuned} in the
setting of Section~\ref{sec:model}. Like in the case of MS, for the purpose of this paper the set of
possible thresholds $t$ is restricted to $\{0.05,\,0.1,\,0.15, \ldots, 0.9, \, 0.95\}$.

\textbf{Tuning APCC for APCCv.} Similarly to \eqref{eq:ACC.tuned}, the idea is to minimise the variance of the 
estimator under $\mathrm{Q}$ while
controlling the size of the denominator in \eqref{eq:APCC}. For $0 < \pi < 1$ define
\begin{equation*}
h_{\pi}(x) \ = \ \frac{\pi\,f^+(x)}{\pi\,f^+(x)+(1-\pi)\,f^-(x)},
\end{equation*}
where $f^+$ and $f^-$ are the class-conditional densities of the features. Then it holds that
\begin{equation*}
h_{\mathrm{P}[Y=1]}(x) \ = \ \mathrm{P}[Y=1\,|\,X](x).
\end{equation*}
A good choice for $\pi$ could be $\pi^\ast$ with
\begin{equation}\label{eq:APCC.tuned}
\pi^\ast \ = \ \arg \min\limits_{0 < \pi < 1} \frac{\mathrm{var}_\mathrm{Q}[h_{\pi}(X)]}
            {\bigl(\mathrm{E}_\mathrm{P}[h_{\pi}(X)\,|\,Y=1]-\mathrm{E}_\mathrm{P}[h_{\pi}(X)\,|\,Y=-1]\bigr)^2}.
\end{equation}
For the purpose of this paper the set of
possible parameters $\pi$  in \eqref{eq:APCC.tuned} is restricted to $\{0.05,\,0.1,\,0.15,$ $\ldots, 0.9, \, 0.95\}$.
In the setting of Section~\ref{sec:model}, let $a$ be defined as in \eqref{eq:trainCondProb} and let
\begin{equation*}
b_\pi \ = \ \frac{\nu^2-\mu^2}{2\,\sigma^2} + \log\left(\frac{1-\pi}{\pi}\right).
\end{equation*}
Then, analogously to \eqref{eq:APCC.here}, in the
setting of Section~\ref{sec:model} the following formulae are obtained for use in the 
calculations of this paper for \eqref{eq:APCC.tuned}:
\begin{align}
\frac{1}{n} \sum_{i=1}^n h_\pi(x_{i, \mathrm{Q}}) &\ =\ 
\frac{1}{n} \sum_{i=1}^n \frac{1}{1 + \exp(a\,x_{i, \mathrm{Q}} + b_\pi)},\notag\\
\mathrm{E}_\mathrm{P}\bigl[h_\pi(X)\,|\,Y=-1] &\ =\ 1 - \int_{-\infty}^\infty \frac{\varphi_{\mu, \sigma}
(x_{i, \mathrm{Q}})}{1 + \exp(a\,x_{i, \mathrm{Q}} + b_\pi)}\, d\,x, \\
\mathrm{E}_\mathrm{P}\bigl[h_\pi(X)\,|\,Y=1\bigr] &\ =\ 1 - \int_{-\infty}^\infty \frac{\varphi_{\nu, \sigma}
(x_{i, \mathrm{Q}})}{1 + \exp(a\,x_{i, \mathrm{Q}} + b_\pi)}\, d\,x.\notag
\end{align}

%%%
% New section
%%%

\subsection{Prevalence estimation by distance minimisation}
\label{sec:DistanceEst}

The idea for prevalence estimation by distance minimisation is to obtain an 
estimate $\widehat{q}$ of $\mathrm{Q}[Y=1] = q$ by solving the following optimisation problem:
\begin{equation}\label{eq:DistanceMin}
\widehat{q}\ = \ \arg \min\limits_{0\le q \le 1} d\left(\mathrm{Q}(X), q\,\mathrm{P}[X \in \cdot\,|\,Y=1] +
                        (1-q)\,\mathrm{P}[X \in \cdot\,|\,Y=-1]\right).
\end{equation}
Here $d$ denotes a distance measure of probability measures with the following two properties:
\begin{enumerate}
\item $d(M_1, M_2) \ge 0$ for all probability measures $M_1$, $M_2$ to which $d$ is applicable.
\item $d(M_1, M_2) = 0$ if and only if $M_1 = M_2$.
\end{enumerate}
There is no need for $d$ to be a metric (i.e.\ asymmetric distance measures $d$ 
with $d(M_1, M_2) \not= d(M_2, M_1)$ for some $M_1$, $M_2$ are permitted). By property 2),
distance minimisation estimators defined by \eqref{eq:DistanceMin} are Fisher consistent for
estimating the positive class prevalence of the test population under prior probability shift.
In the following subsections three approaches to prevalence estimation 
based on distance minimisation are introduced that have been suggested in the literature 
and appear to be popular.

%%%
% New section
%%%

\subsubsection{Prevalence estimation by minimising the Hellinger distance}
\label{sec:Hellinger}

The Hellinger distance\footnote{%
See \citet{gonzalez2013class} and \citet{castano.analisis} for more information on the Hellinger distance approach
to prevalence estimation.} $d_H$ of two probability measures $M_1$, $M_2$ on the same domain is defined in measure-theoretic terms by
\begin{equation}\label{eq:Hellinger}
d_H(M_1, M_2) \ = \ \frac 1 2 \int \left(\sqrt{\frac{d\,M_1}{d\,\lambda}} - \sqrt{\frac{d\,M_2}{d\,\lambda}}\right)^2\, d \lambda,
\end{equation}
where $\lambda$ is any measure on the same domain such that both $M_1$ and $M_2$ are absolutely continuous with respect to 
$\lambda$. The value of $d_H(M_1, M_2)$ does not depend upon the choice of $\lambda$.

In practice, the calculation of the Hellinger distance must take into account that most of time it has to be estimated
from sample data. Therefore, the right-hand side of \eqref{eq:Hellinger} is discretized by 
(in the setting of Section~\ref{sec:setting})
decomposing the feature space $\mathcal{X}$ into a finite number of subsets or bins $\mathcal{X}_1, \ldots, \mathcal{X}_b$ and
evaluating the probability measures whose distance is to be measured on these bins. This leads to the following 
approximative version of the minimisation problem \eqref{eq:DistanceMin}:
\begin{equation}\label{eq:Hellinger.bins}
   \widehat{q}\ = \ \arg \min\limits_{0\le q \le 1} \sum_{i=1}^b \left(\sqrt{Q[X\in\mathcal{X}_i]} - 
   \sqrt{q\,\mathrm{P}[X \in \mathcal{X}_i\,|\,Y=1] +
                        (1-q)\,\mathrm{P}[X \in \mathcal{X}_i\,|\,Y=-1]}\right)^2.
\end{equation}
If the feature space $\mathcal{X}$ is multi-dimensional, e.g.\ $\mathcal{X} \subset \mathbb{R}^d$ for some $d \ge 2$, 
\citet{gonzalez2013class} also suggest minimising the Hellinger distance 
separately across all the $d$ dimensions of the feature
vector $X=(X_1, \ldots, X_d)$. In this case, the feature space 
$\mathcal{X} = \mathcal{X}_1 \times \ldots \times \mathcal{X}_d$ is decomposed component-wise in $b$ bins 
and 
\eqref{eq:Hellinger.bins} is modified to become
\begin{equation*}
  \widehat{q}\ = \ \arg \min\limits_{0\le q \le 1} \sum_{k=1}^d \sum_{i=1}^b \left(\sqrt{Q[X_k\in\mathcal{X}_{i,k}]} - 
   \sqrt{q\,\mathrm{P}[X_k\in\mathcal{X}_{i,k}\,|\,Y=1] +
                        (1-q)\,\mathrm{P}[X_k\in\mathcal{X}_{i,k}\,|\,Y=-1]}\right)^2,
\end{equation*}
where $\mathcal{X}_k = \bigcup_{i=1}^b \mathcal{X}_{i,k}$.
For the purposes of this paper, \eqref{eq:Hellinger.bins} has been adapted to become
\begin{equation}\label{eq:semiHellinger}
\widehat{q}\ = \ \arg \min\limits_{0\le q \le 1} \sum_{i=1}^b 
\left(\sqrt{\frac{1}{n}\sum_{j=1}^n I\left(x_{j, \mathrm{Q}}\in\mathcal{X}_i\right)} - 
   \sqrt{q\,\mathrm{P}[X \in \mathcal{X}_i\,|\,Y=1] +
                        (1-q)\,\mathrm{P}[X \in \mathcal{X}_i\,|\,Y=-1]}\right)^2,
\end{equation}
where $(x_{1, \mathrm{Q}},$ $\ldots$ $x_{n, \mathrm{Q}})$ $\in \mathbb{R}^n$ is a sample of
features of instances generated under the test population distribution $\mathrm{Q}(X)$ and the $\mathrm{P}$-terms
must be estimated from the training sample if it is finite and can be exactly pre-calculated in
the case of an infinite training sample. In the latter case,
\eqref{eq:semiHellinger} has to be modified to reflect the binormal setting of Section~\ref{sec:model} for the
training population distribution:
\begin{subequations}
\begin{multline}\label{eq:semiHellinger1}
\widehat{q}\ = \ \arg \min\limits_{0\le q \le 1} \sum_{i=1}^b 
\left(\sqrt{\frac{1}{n}\sum_{j=1}^n I\left(\ell_{i-1} < x_{j, \mathrm{Q}} \le \ell_i\right)} \ - \right.\\ 
   \left.\sqrt{q\,\left(\Phi\left(\frac{\ell_i-\nu}{\sigma}\right)-\Phi\left(\frac{\ell_{i-1}-\nu}{\sigma}\right)\right) +
   (1-q)\,\left(\Phi\left(\frac{\ell_i-\mu}{\sigma}\right)-\Phi\left(\frac{\ell_{i-1}-\mu}{\sigma}\right)\right)}\ \right)^2,
\end{multline}
if $\mathcal{X}_i = [\ell_{i-1}, \ell_i)$ for $-\infty = \ell_0 < \ell_1 < \ldots \ \ell_{b-1} < \ell_b = \infty$. 
For this paper, the number $b$ of bins\footnote{%
See \citet{Maletzke&etAl.Quantification} for critical comments regarding the choice of the number of bins.} in \eqref{eq:semiHellinger1} has been chosen to be 4 or 8, and
the boundaries of the bins have been defined as follows\footnote{$\Phi^{-1}$ is
the inverse function to the standard normal distribution function.}:
\begin{equation}
\ell_i \ =\ \sigma\,\Phi^{-1}\left(\frac{i}{b}\right) + \frac{\mu + \nu}{2}, \qquad i = 1, \ldots, b-1.
\end{equation}
\end{subequations}

%%%
% New section
%%%

\subsubsection{Prevalence estimation by minimising the Energy distance}
\label{sec:Energy}

\citet{kawakubo2016computationally} and \citet{castano.analisis} provide background 
information for the application of the Energy distance approach to prevalence estimation.

Denote by $V$ and $V'$ respectively the projection on the first $d$ components 
and the last $d$ components respectively of  $\mathbb{R}^{2\,d}$, i.e.
\begin{equation*}
V(x) \ = \ (x_1, \ldots, x_d)\quad\text{and}\quad V'(x) \ = \ (x_{d+1}, \ldots, x_{2\,d}), 
\quad \text{for}\ x \in \mathbb{R}^{2\,d}.
\end{equation*}
$V$ and $V'$ are also used to denote the identity mapping on $\mathbb{R}^d$, i.e.\ $V(x) = x$ and 
$V'(x) = x$ for $x \in \mathbb{R}^d$.

Let $M_1$, $M_2$ be two probability measures on $\mathbb{R}^d$. 
Then $M_1 \otimes M_2$ denotes the product measure of $M_1$ and $M_2$ on $\mathbb{R}^{2\,d}$. 
Hence $M_1 \otimes M_2$ is
the probability measure on $\mathbb{R}^{2\,d}$ such that $V$ and $V'$ are stochastically independent under $M_1 \otimes M_2$.

Denote by $||x||$ the Euclidean norm of $x \in \mathbb{R}^d$. Then the Energy distance $d_E$ of two probability 
measures $M_1$, $M_2$ on $\mathbb{R}^d$  with $\mathrm{E}_{M_1}[||V||^2]<\infty$ and 
$\mathrm{E}_{M_2}[||V'||^2]<\infty$ can be represented as
\begin{equation}\label{eq:Energy}
d_E(M_1, M_2) \ = \ 2\,\mathrm{E}_{M_1 \otimes M_2}\bigl[||V-V'||\bigr] - \mathrm{E}_{M_1 \otimes M_1}\bigl[||V-V'||\bigr] -
                \mathrm{E}_{M_2 \otimes M_2}\bigl[||V-V'||\bigr].
\end{equation}
Recall that in this section the aim is to estimate class prevalences by solving the optimisation problem 
\eqref{eq:DistanceMin}. To do so by means of minimising the Energy distance, fix a function $h: \mathcal{X} \to \mathbb{R}$
and choose as probability measures $M_1$ and $M_2$ the distributions of $h(X)$ under the probability measures
$\mathrm{Q}$ and $q\,\mathrm{P}[X \in \cdot\,|\,Y=1]$ $+\, (1-q)\,\mathrm{P}[X \in \cdot\,|\,Y=-1]$ whose distance is
minimised in \eqref{eq:DistanceMin}:
\begin{equation*}
\begin{split}
M_1(D) & \ = \ \mathrm{Q}[h(X) \in D],\\
M_2(D)  & \ = \ q\,\mathrm{P}[h(X) \in D\,|\,Y=1] + (1-q)\,\mathrm{P}[h(X) \in D\,|\,Y=-1],
\end{split}
\end{equation*}
for $0 \le q \le 1$ and $D\subset \mathbb{R}$ such that all involved probabilities are well-defined. With
this choice for $M_1$ and $M_2$, it follows from \eqref{eq:Energy} that
\begin{multline}\label{eq:EnergyQuant}
d_E(M_1, M_2) \ = \ 2\,q\,\mathrm{E}_{\mathrm{Q} \otimes \mathrm{P}^+}\bigl[|h(V)-h(V')|\bigr]
+ 2\,(1-q)\,\mathrm{E}_{\mathrm{Q} \otimes \mathrm{P}^-}\bigl[|h(V)-h(V')|\bigr] \\
- \mathrm{E}_{\mathrm{Q} \otimes \mathrm{Q}}\bigl[|h(V)-h(V')|\bigr]
- q^2\,\mathrm{E}_{\mathrm{P}^+ \otimes \mathrm{P}^+}\bigl[|h(V)-h(V')|\bigr]
- (1-q)^2\,\mathrm{E}_{\mathrm{P}^- \otimes \mathrm{P}^-}\bigl[|h(V)-h(V')|\bigr]\\
- 2\,q\,(1-q)\,\mathrm{E}_{\mathrm{P}^+ \otimes \mathrm{P}^-}\bigl[|h(V)-h(V')|\bigr],
\end{multline}
with $\mathrm{P}^+ = \mathrm{P}[X \in \cdot\,|\,Y=1]$ and $\mathrm{P}^- = \mathrm{P}[X \in \cdot\,|\,Y=-1]$.
The unique minimising value $\widehat{q}$ of $q$ for the right-hand side of \eqref{eq:EnergyQuant} is found to be
\begin{align}
\widehat{q} & = \frac{A}{B}, \quad \text{with}\label{eq:q.energy}\\
A & =
\mathrm{E}_{\mathrm{Q} \otimes \mathrm{P}^-}[|h(V)-h(V')|] - 
\mathrm{E}_{\mathrm{Q} \otimes \mathrm{P}^+}[|h(V)-h(V')|]\notag\\
& \qquad\qquad - \mathrm{E}_{\mathrm{P}^-\otimes \mathrm{P}^-}[|h(V)-h(V')|] +
\mathrm{E}_{\mathrm{P}^+ \otimes \mathrm{P}^-}[|h(V)-h(V')|], \notag\\
B & = 2\,\mathrm{E}_{\mathrm{P}^+ \otimes \mathrm{P}^-}[|h(V)-h(V')|] -
\mathrm{E}_{\mathrm{P}^- \otimes \mathrm{P}^-}[|h(V)-h(V')|] -
\mathrm{E}_{\mathrm{P}^+ \otimes \mathrm{P}^+}[|h(V)-h(V')|].\notag
\end{align}
The fact that there is a closed-form solution for the estimate $\widehat{q}$ in the two-class case is one
of the advantages of the Energy distance approach.

When both the training and the test samples are finite, all the sub-terms of $A$ and $B$ in 
\eqref{eq:q.energy} can be empirically estimated in a straight-forward way.
For the semi-finite version of \eqref{eq:q.energy}, when the training sample is infinite,
the terms $\mathrm{E}_{\mathrm{Q} \otimes \mathrm{P}^-}[|h(V)-h(V')|]$ and
$\mathrm{E}_{\mathrm{Q} \otimes \mathrm{P}^+}[|h(V)-h(V')|]$ have to be replaced by empirical approximations
based on a sample $(x_{1, \mathrm{Q}},$ $\ldots$ $x_{n, \mathrm{Q}})$ $\in \mathbb{R}^n$ of
features of instances generated under the test population distribution $\mathrm{Q}(X)$, while the
population measures for $\mathrm{P}^-$ and $\mathrm{P}^+$ respectively are kept:
\begin{equation} \label{eq:q.emp}
\begin{split}
\mathrm{E}_{\mathrm{Q} \otimes \mathrm{P}^-}[|h(V)-h(V')|] &\ \approx\ 
\mathrm{E}_{\mathrm{P}^-}\left[ \frac{1}{n}\sum_{j=1}^n |h(x_{j, \mathrm{Q}}) - h(V)|\right], \\
\mathrm{E}_{\mathrm{Q} \otimes \mathrm{P}^+}[|h(V)-h(V')|] &\ \approx\ 
\mathrm{E}_{\mathrm{P}^+}\left[ \frac{1}{n}\sum_{j=1}^n |h(x_{j, \mathrm{Q}}) - h(V)|\right].
\end{split}
\end{equation}
Accordingly, for the case where in the setting of Section~\ref{sec:model} with the double 
binormal model the training sample is infinite, 
the following formulae have been coded for \eqref{eq:q.emp} and 
the terms $A$ and $B$ in \eqref{eq:q.energy}, with 
$h(x) = \mathrm{P}[Y=1\,|\,X](x) = \frac{1}{1 + \exp(a\,x + b)}$ (parameters $a, b$ as 
in \eqref{eq:trainCondProb}):
\begin{align}
\mathrm{E}_{\mathrm{P}^-}\left[ \frac{1}{n}\sum_{j=1}^n |h(x_{j, \mathrm{Q}}) - h(V)|\right] & =
\frac{1}{n} \int_{-\infty}^\infty \varphi_{\mu, \sigma}(x)\,\sum_{j=1}^n \bigm| \frac{1}{1 + \exp(a\,x + b)} -
\frac{1}{1 + \exp(a\,x_{j, \mathrm{Q}} + b)}\bigm|\,d\,x,\notag
\\
\mathrm{E}_{\mathrm{P}^+}\left[ \frac{1}{n}\sum_{j=1}^n |h(x_{j, \mathrm{Q}}) - h(V)|\right] & = 
\frac{1}{n} \int_{-\infty}^\infty \varphi_{\nu, \sigma}(x)\,\sum_{j=1}^n \bigm| \frac{1}{1 + \exp(a\,x + b)} -
\frac{1}{1 + \exp(a\,x_{j, \mathrm{Q}} + b)}\bigm|\,d\,x,\notag
\\
\mathrm{E}_{\mathrm{P}^-\otimes \mathrm{P}^-}[|h(V)-h(V')|]  & =
4 \int_{-\infty}^\infty \frac{\varphi_{\mu, \sigma}(x)\,\Phi_{\mu, \sigma}(x)}
    {1 + \exp(a\,x + b)}\, d\,x - 2 \int_{-\infty}^\infty \frac{\varphi_{\mu, \sigma}(x)}
    {1 + \exp(a\,x + b)}\, d\,x,
\\
\mathrm{E}_{\mathrm{P}^+ \otimes \mathrm{P}^-}[|h(V)-h(V')|] & = 
2 \int_{-\infty}^\infty \frac{\varphi_{\mu, \sigma}(x)\,\Phi_{\nu, \sigma}(x)}
    {1 + \exp(a\,x + b)}\, d\,x - \int_{-\infty}^\infty \frac{\varphi_{\mu, \sigma}(x)}
    {1 + \exp(a\,x + b)}\, d\,x\ +\notag\\
& \qquad 2 \int_{-\infty}^\infty \frac{\varphi_{\nu, \sigma}(x)\,\Phi_{\mu, \sigma}(x)}
    {1 + \exp(a\,x + b)}\, d\,x - \int_{-\infty}^\infty \frac{\varphi_{\nu, \sigma}(x)}
    {1 + \exp(a\,x + b)}\, d\,x, \notag   
\\
\mathrm{E}_{\mathrm{P}^+ \otimes \mathrm{P}^+}[|h(V)-h(V')|] & =
4 \int_{-\infty}^\infty \frac{\varphi_{\nu, \sigma}(x)\,\Phi_{\nu, \sigma}(x)}
    {1 + \exp(a\,x + b)}\, d\,x - 2 \int_{-\infty}^\infty \frac{\varphi_{\nu, \sigma}(x)}
    {1 + \exp(a\,x + b)}\, d\,x.\notag
\end{align}

%%%
% New section
%%%

\subsubsection{Prevalence estimation by minimising the Kullback-Leibler distance}
\label{sec:KL}

In this section, the approach to prevalence estimation 
based on minimising the Kullback-Leibler distance\footnote{%
Kullback-Leibler distance is also called Kullback-Leibler divergence \citep[for instance in][]{duPlessis2014110}.} 
is presented. In practical applications, this approach
is equivalent to maximum likelihood estimation of the class prevalences. Due to the asymptotic 
efficiency of maximum likelihood estimators in terms of the variances of the estimates, this approach can serve
as an absolute benchmark for what can be achieved in terms of short confidence intervals.
\citet{saerens2002adjusting} made the EM-algorithm version of the approach -- as a possibility to obtain
the maximum likelihood estimate \citep{redner1984mixture} -- popular in the machine learning community.

Let $M_1$, $M_2$ be two probability measures on the same domain. Assume that both $M_1$ and $M_2$
are absolutely continuous with respect to a measure $\lambda$ on the same domain, with densities 
\begin{equation*}
f_1 \ = \ \frac{d\,M_1}{d\,\lambda}\qquad \text{and} \qquad f_2 \ = \ \frac{d\,M_2}{d\,\lambda}
\end{equation*}
respectively. If the densities $f_1$, $f_2$ are positive, the Kullback-Leibler distance of $M_2$ to $M_1$ then is defined as
\begin{equation}\label{eq:KL}
d_{KL}(M_2||M_1) \ = \ \int f_1\,\log\left(\frac{f_1}{f_2}\right)\,d \lambda.
\end{equation}
In contrast to the Hellinger and Energy distances which were introduced in sections~\ref{sec:Hellinger} and 
\ref{sec:Energy} respectively, 
the Kullback-Leibler distance is not symmetric in its arguments, i.e.\ in general
$d_{KL}(M_2||M_1) \not= d_{KL}(M_1||M_2)$ may occur. In \eqref{eq:KL}
the lack of symmetry is indicated by separating  the arguments not by a comma but by the sign $||$. But while 
$d_{KL}$ is not a metric it still has the properties $d_{KL}(M_2||M_1) \ge 0$ and $d_{KL}(M_2||M_1) = 0$
if and only if $M_1 = M_2$.

The choice $M_1 = \mathrm{Q}(X)$ and $M_2 = q\,\mathrm{P}[X \in \cdot\,|\,Y=1] + (1-q)\,\mathrm{P}[X \in \cdot\,|\,Y=-1]$
leads to a computationally convenient Kullback-Leibler version of the minimisation problem \eqref{eq:DistanceMin}. 
One has to assume that there are a measure $\lambda$ on the feature space $\mathcal{X}$ and densities $f_\mathrm{Q}$, $f^+$ and
$f^-$ such that
\begin{equation*}%\label{eq:densities}
\begin{split}
f_\mathrm{Q}(X) &\ =\ \frac{d\,\mathrm{Q}(X)}{d\,\lambda}\ >\ 0, \\
f^+(X) &\ =\ \frac{d\,\mathrm{P}[X \in \cdot\,|\,Y=1]}{d\,\lambda},\\ 
f^-(X) &\ =\ \frac{d\,\mathrm{P}[X \in \cdot\,|\,Y=-1]}{d\,\lambda}, \quad\text{and}\\
f^+(X) + f^-(X) &\ >\ 0
\end{split}
\end{equation*}
This gives the following optimisation problem:
\begin{align}
\widehat{q}& \ = \ \arg\min\limits_{0\le q\le 1} 
    \int f_\mathrm{Q}\,\log\left(\frac{f_\mathrm{Q}(X)}{q\,f^+(X) + (1-q)\,f^-(X)}\right)\,d \lambda \notag\\
 & \ = \ \arg\max\limits_{0\le q\le 1} 
    \mathrm{E}_\mathrm{Q}\left[\log\left(q\,f^+(X) + (1-q)\,f^-(X)\right)\right] \notag \\
    &\ = \ \arg\max\limits_{0\le q\le 1} 
    \mathrm{E}_\mathrm{Q}\left[\log\left(q\, (R(X)-1) + 1\right)\right],\label{eq:EMmax}
\end{align}
where the density ratio $R(X)$ is defined by
\begin{equation*}%\label{eq:densratio}
R(X)\ =\ \frac{f^+(X)}{f^-(X)}
 \end{equation*} 
and additionally it must hold that $f^-(X) > 0$. 
Under fairly general smoothness conditions, 
the right-hand side of \eqref{eq:EMmax} can be differentiated with respect to $q$. This gives
the following necessary condition for optimality in \eqref{eq:EMmax}:
\begin{equation}\label{eq:REq}
0 \ = \ \mathrm{E}_\mathrm{Q}\left[\frac{R(X)-1}{\widehat{q}\, (R(X)-1) + 1}\right].
\end{equation}
When both the training and the test samples are finite, the right-hand side of \eqref{eq:REq} 
as a function of $\widehat{q}$ can be empirically estimated in a straight-forward way.

For the semi-finite setting according to Section~\ref{sec:model} with infinite training sample, 
it can be assumed that the density ratio $R$ 
is fully known by the specification of the model. 
In contrast, the test population distribution $\mathrm{Q}(X)$ of the features is
only known through a sample (or empirical distribution) $x_{1, \mathrm{Q}}, \ldots,$ 
$x_{n, \mathrm{Q}}$ that was sampled from
$\mathrm{Q}(X)$. Replacing the expectation with respect to $\mathrm{Q}$ in \eqref{eq:REq} 
with a sample average gives the equation
\begin{subequations}
\begin{equation}\label{eq:maxlik}
0 \ = \ \sum_{i=1}^n \frac{R(x_{i, \mathrm{Q}})-1}{\widehat{q}\, (R(x_{i, \mathrm{Q}})-1) + 1}
\end{equation}
as an approximative necessary condition for  $\widehat{q}$\/ 
to minimise the Kullback-Leibler distance in \eqref{eq:EMmax}. Solving \eqref{eq:maxlik} results 
in an approximation $\widehat{q}_n$ of $\widehat{q}$ and therefore $\mathrm{Q}[Y=1]$.

It is not hard to see \citep[see, for instance, Lemma~4.1 of][]{tasche2013law} that \eqref{eq:maxlik} has a
unique solution $\widehat{q}_n$ with $0 \le \widehat{q}_n\le 1$ if and only if
\begin{equation}\label{eq:cond}
R(x_{i, \mathrm{Q}}) \not=1 \ \text{for at least one}\ i,\quad \frac 1 n
\sum_{i=1}^n R(x_{i, \mathrm{Q}}) \ge 1 \quad \text{and}\quad \frac 1 n \sum_{i=1}^n \frac1{R(x_{i, \mathrm{Q}})} \ge 1.
\end{equation}
\end{subequations}
If \eqref{eq:cond} holds the solution of \eqref{eq:maxlik} is $\widehat{q}_n =0$ if and only if 
$\frac 1 n \sum_{i=1}^n R(x_{i, \mathrm{Q}}) = 1$ and the solution is $\widehat{q}_n =1$ if and only if
$\frac 1 n \sum_{i=1}^n\frac1{R(x_{i, \mathrm{Q}})} = 1$. For the purpose of this paper, clipping\footnote{%
In the literature on prevalence estimation, clipping is applied routinely 
\citep[see, for instance,][]{forman2008quantifying}. In general, one should be careful with clipping because
the fact that there is no estimate between 0 and 1 could be a sign that the assumption of prior
probability shift is violated. This is not an issue in the setting of this paper because prior 
probability shift is created by the design of the simulation study.} is applied when solving \eqref{eq:maxlik}.
Since, as shown in the proof of Lemma~4.1 of \citet{tasche2013law}, it is not possible that both
$\frac 1 n\sum_{i=1}^n R(x_{i, \mathrm{Q}}) < 1$ and 
$\frac 1 n\sum_{i=1}^n \frac1{R(x_{i, \mathrm{Q}})} < 1$ occur, it makes sense 
to set
\begin{itemize}
\item $\widehat{q}_n = 0$ if $\frac 1 n\sum_{i=1}^n R(x_{i, \mathrm{Q}}) \le 1$ and
\item $\widehat{q}_n = 1$ if $\frac 1 n\sum_{i=1}^n \frac1{R(x_{i, \mathrm{Q}})} \le 1$.
\end{itemize}
Equation~\eqref{eq:maxlik} happens also to be the first order condition for the maximum likelihood estimator 
of the test population prevalence of the positive class $\mathrm{Q}[Y=1]$, see \citet{peters1976numerical}.
A popular method to determine the maximum likelihood estimates for the class prevalences in mixture proportion
problems like the one of this paper is to deploy an Expectation Maximisation (EM) approach 
\citep{redner1984mixture, saerens2002adjusting}.
However, in the specific semi-finite context of this paper, and more generally in the two-classes case, it
is more efficient to solve \eqref{eq:maxlik} directly by an appropriate numerical algorithm (see for instance
the documentation of the R-function `uniroot').

The confidence and prediction intervals based on $\widehat{q}$, as found by solving the empirical version of
\eqref{eq:REq} or \eqref{eq:maxlik},
are determined with the following two methods:
\begin{itemize}
\item Method MLboot: Bootstrapping the samples $x_{1, \mathrm{P}^+}$, $\ldots$, $x_{{m^+}, \mathrm{P}^+}$,  
$x_{{1}, \mathrm{P}^-}$, $\ldots$, $x_{{m^-}, \mathrm{P}^-}$, and 
$x_{1, \mathrm{Q}}$, $\ldots$, $x_{{n}, \mathrm{Q}}$ from Section~\ref{sec:calc} 
and creating a sample of $\widehat{q}_n$ by solving \eqref{eq:maxlik}
for each of the bootstrapping samples.
\item Method MLinf: Asymptotic approximation by making use of the central limit theorem for maximum likelihood estimators 
\citetext{see, e.g., Theorem~10.1.12 of \citealp{Casella&Berger}}. According to this limit theorem,  
$\sqrt{n}\,(\widehat{q}_n - \mathrm{Q}[Y=1])$ converges for $n\to\infty$ toward a normal distribution
with mean 0 and variance $v$ given by 
\begin{subequations}
\begin{equation}\label{eq:asymVar}
v \ = \ \frac{1}{\displaystyle\mathrm{E}_\mathrm{Q}\left[ \left(\frac{R(X)-1}{q\, (R(X)-1) + 1}\right)^2\right]},
\end{equation}
where $q$ is the true positive class prevalence of the population underlying the test sample.
The right-hand side of \eqref{eq:asymVar} is approximated 
with an estimate $v_n$ that is based on a sample average:
\begin{equation}
v_n \ = \ \dfrac{n}{\displaystyle\sum_{i=1}^n \left(\frac{R(x_{i, 
\mathrm{Q}})-1}{\widehat{q}_n\, (R(x_{i, \mathrm{Q}})-1) + 1}\right)^2},
\end{equation}
where $\widehat{q}_n$ denotes the -- unique if any -- solution of \eqref{eq:maxlik} in the unit interval $[0,1]$ 
and $x_{1, \mathrm{Q}}, \ldots, x_{n, \mathrm{Q}}$ was generated under the test population distribution $\mathrm{Q}(X)$. In the
binormal setting of Section~\ref{sec:model}, the density ratio $R(x)$ is given by \eqref{eq:ratioBinorm} if the
training sample is infinite and can be derived from the posterior class probabilities obtained by logistic regression
if the training sample is finite.
\end{subequations}
\end{itemize}

%%%
% New section
%%%

\section{Appendix: Analysis of Error Adjusted Bootstrapping}
\label{sec:appB} 

Without mentioning  
explicitly the notion of prediction intervals, \citet{daughton2019constructing} 
considered the problem of how to construct prediction intervals for the realised positive class
prevalence with correct coverage rates. Their `error adjusted bootstrapping' approach works for crisp classifiers only.

Let $g$ be a crisp classifier as defined in Section~\ref{sec:ACCetRel}. In the notation of that section then
it holds that
\begin{equation}\label{eq:daughton}
\mathrm{Q}[Y=1] \ = \ \mathrm{Q}[Y=1\,|\,g(X)=1]\,\mathrm{Q}[g(X)=1] + 
        \mathrm{Q}[Y=1\,|\,g(X)=-1]\,\mathrm{Q}[g(X)=-1].
\end{equation}
Hence the event `an instance in the test sample turns out to have a positive label' can
be simulated in three steps:
\begin{enumerate}
\item Apply the classifier $g$ to the bootstrapped features of an instance in the test sample.
\item If a positive label is predicted by $g$, simulate a Bernoulli variable with success probability
$\mathrm{Q}[Y=1\,|\,g(X)=1]$. If a negative label is predicted by $g$, simulate a Bernoulli variable with success probability
$\mathrm{Q}[Y=1\,|\,g(X)=-1]$. 
\item In both cases, if the outcome of the Bernoulli variable is success, count the result as positive class, otherwise
as negative class.
\end{enumerate}
By \eqref{eq:daughton}, the probability of the positive class in this experiment is the prevalence of the
positive class in the test population distribution. Repeat the experiment for all the instances
in the bootstrapped test sample. Then the relative frequency of the positive outcomes of the experiments is
an approximate realisation of the relative frequency of the positive class labels in the test sample which in the same way 
as by the binomial approach described in Section~\ref{sec:calc} can be used to construct a bootstrap prediction interval.

\citet{daughton2019constructing} noted that this approach worked if the `predictive values' $\mathrm{Q}[Y=1\,|\,g(X)=1]$
and $\mathrm{Q}[Y=1\,|\,g(X)=-1]$ of the test sample ($\mathrm{Q}[Y=1\,|\,g(X)=1]$ is also called `precision' in
the literature) were the same as in the training sample and hence could be estimated in the training sample:
\begin{equation}\label{eq:precision}
\mathrm{Q}[Y=1\,|\,g(X)=1] \ = \ \mathrm{P}[Y=1\,|\,g(X)=1], \qquad
\mathrm{Q}[Y=1\,|\,g(X)=-1] \ = \ \mathrm{P}[Y=1\,|\,g(X)=-1].
\end{equation}
Unfortunately, \eqref{eq:precision} does not hold under prior probability shift as is implied by the following
representation of the precision in terms of $TPR = \mathrm{P}[g(X)=1\,|\,Y=1]$, 
$FPR = \mathrm{P}[g(X)=1\,|\,Y=-1]$ and test population prevalence $p$:
\begin{equation}\label{eq:precisionRep}
\mathrm{P}[Y=1\,|\,g(X)=1] \ = \ \frac{p\,TPR}{p\,(TPR-FPR) + FPR}.
\end{equation}
Under prior probability shift, TPR and FPR are not changed, but $p$ changes. Hence, by replacing $p$ with $q\not=p$ 
on the right-hand side of \eqref{eq:precisionRep}, it follows that
$$\mathrm{Q}[Y=1\,|\,g(X)=1] \ = \ \frac{q\,TPR}{q\,(TPR-FPR) + FPR}\ \not= \ \mathrm{P}[Y=1\,|\,g(X)=1].$$
Therefore, under prior probability shift, the approach by \citet{daughton2019constructing} is unlikely to work
in general. See Table~\ref{tab:DnP} in Section~\ref{sec:predVSconf} for a numerical example. 
It is not clear if requiring that \eqref{eq:precision} holds results in defining an instance 
of data set shift which might occur in the real world.

\end{document}